\documentclass{article}

\usepackage[final]{neurips_data_2023}
\usepackage{graphicx}
\graphicspath{{./materials/}}
\usepackage{CJKutf8}

\usepackage[utf8]{inputenc} % allow utf-8 input
\usepackage[T1]{fontenc}    % use 8-bit T1 fonts
\usepackage{hyperref}       % hyperlinks
\usepackage{url}            % simple URL typesetting
\usepackage{booktabs}       % professional-quality tables
\usepackage{amsfonts}       % blackboard math symbols
\usepackage{nicefrac}       % compact symbols for 1/2, etc.
\usepackage{microtype}      % microtypography
\usepackage{xcolor}         % colors
\usepackage{graphicx}
\usepackage{booktabs}
\usepackage{multirow}
\usepackage{makecell}
\usepackage{tabularx}
\usepackage{wrapfig}
\usepackage{caption}
\usepackage{enumitem}
\usepackage{float}
\newcommand{\myparagraph}[1]{\textbf{#1}\hspace{1.8ex}}

\title{Benchmarking Large Language Models on CMExam - A Comprehensive Chinese Medical Exam Dataset}
\author{%
  Junling Liu$^{1\dagger}$\thanks{Corresponding Author. $^{\dagger}$Co-first authors}\quad
  Peilin Zhou$^{2\dagger}$  \quad
  Yining Hua$^{3,4\dagger}$ \quad
  Dading Chong$^{5}$ 
  \\
  \textbf{Zhongyu Tian}$^{6}$ \quad
  \textbf{Andrew Liu}$^{5}$ \quad
  \textbf{Helin Wang}$^{7}$ \quad
  \textbf{Chenyu You}$^{8}$ 
  \\
   \textbf{Zhenhua Guo}$^{9}$ \quad
   \textbf{Lei Zhu}$^{10}$ \quad
   \textbf{Michael Lingzhi Li}$^{4,11}$
   \vspace{1mm}
  \\
$^{1}$Alibaba Group
$^{2}$Hong Kong University of Science and Technology (Guangzhou)\vspace{0.5mm}\\
$^{3}$Harvard University
$^{4}$Boston Children's Hospital
$^{5}$Peking University \vspace{0.5mm}\\
$^{6}$Second Affiliated Hospital of Zhejiang University School of Medicine
$^{7}$Johns Hopkins University\vspace{0.5mm}\\
$^{8}$Yale University
$^{9}$Tianyi Traffic Technology
$^{10}$Ant Group
$^{11}$Harvard Business School\vspace{2mm}\\
  \texttt{\{william.liuj, zhoupalin, andrew.promed, cszguo, zhulei0305\}@gmail.com} \\
  \texttt{1601213984@pku.edu.cn, zhongyutian@zju.edu.cn, hwang258@jhu.edu}\\
  \texttt{yininghua@g.harvard.edu, chenyu.you@yale.edu, mili@hbs.edu}
}

\begin{document}

\maketitle

\begin{abstract}
Recent advancements in large language models (LLMs) have transformed the field of question answering (QA). However, evaluating LLMs in the medical field is challenging due to the lack of standardized and comprehensive datasets. To address this gap, we introduce \textbf{CMExam}, sourced from the \textbf{C}hinese National \textbf{M}edical Licensing \textbf{Exam}ination. CMExam consists of 60K+ multiple-choice questions for standardized and objective evaluations, as well as solution explanations for model reasoning evaluation in an open-ended manner. For in-depth analyses of LLMs, we invited medical professionals to label five additional question-wise annotations, including \textit{disease groups}, \textit{clinical departments}, \textit{medical disciplines}, \textit{areas of competency}, and \textit{question difficulty levels}. 
Alongside the dataset, we further 
% construct a panel of LLM comparisons, including 
% (1) two types of downstream tasks; (2) five different annotations; (3) fifteen representative LLMs and QA algorithms.
conducted thorough experiments with representative LLMs and QA algorithms on CMExam.
The results show that GPT-4 had the best accuracy of 61.6\% and a weighted F1 score of 0.617. These results highlight a great disparity when compared to human accuracy, which stood at 71.6\%. For explanation tasks, while LLMs could generate relevant reasoning and demonstrate improved performance after finetuning, they fall short of a desired standard, indicating ample room for improvement. To the best of our knowledge, CMExam is the first Chinese medical exam dataset to provide comprehensive medical annotations. The experiments and findings of LLM evaluation also provide valuable insights into the challenges and potential solutions in developing Chinese medical QA systems and LLM evaluation pipelines.\footnote[1]{The dataset and relevant code are available at ~\url{https://github.com/williamliujl/CMExam}}
\end{abstract}

\section{Introduction}
\begin{figure}[ht]
  \centering
  \includegraphics[width=1\textwidth]{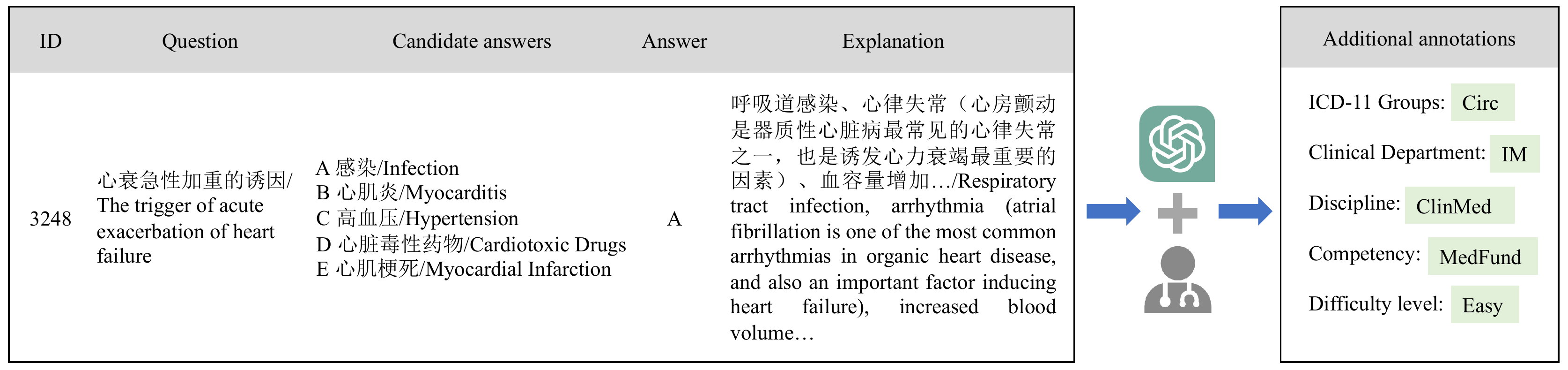}
  \caption{An example question of CMExam. Abbreviations: Circulatory System Diseases (Circ),  Internal Medicine (IM), Clinical Medicine (ClinMed), Medical Fundamentals (MedFund).}
  \label{fig:sample}
\end{figure}
Recent advancements brought by large language models (LLMs) such as T5 \citep{raffel2020exploring} and GPT-4 \citep{OpenAI2023GPT4TR} have revolutionized natural language processing (NLP). However, evaluating LLMs in the medical field poses significant challenges due to the paucity of standardized and comprehensive datasets compiled from reliable and unbiased sources \citep{li2023huatuo}. Most existing medical datasets \citep{hendrycks2020mmlu,abacha2019mediqa,li2023huatuo, peilin2022metscov} for language model evaluation have limitations that hinder comprehensive assessment of LLM performance \citep{nori2023capabilities}. Many datasets are insufficient in terms of size and diversity, preventing a thorough evaluation of LLM capabilities. Furthermore, most datasets primarily focus on text generation tasks rather than utilizing clear choice evaluations, impeding objective and quantitative measurement of LLM performance. Additionally, a majority of these datasets \citep{li2023huatuo,pal2022medmcqa,zhu2020MASH-QA} are sourced from online forums and consumer feedback, which could suffer from significant bias and error.  These challenges are particularly amplified in non-English languages, such as Chinese, due to the pervasive inequality in language resources that exists in the NLP field \citep{bird2020decolonising, zeng2022greenplm, fang2023does}. Overall, due to the lack of qualified evaluation datasets, the strengths and weaknesses of LLMs in the medical field have not been fully studied. 

In response, we present a novel dataset called CMExam to overcome these challenges and benchmark LLM performance. CMExam is sourced from authentic medical licensing exams. It contains more than 60K questions and utilizes the multiple-choice question format to allow standardized and objective evaluations. Questions in CMExam have corresponding solution explanations that can be used to test LLM's reasoning ability in an open-ended manner. To offer diverse perspectives for measuring LLM performance in the medical field, we created five additional question-wise annotation dimensions based on authenticated resources and objective metrics. To reduce the substantial time and labor costs associated with annotating large-scale datasets, we propose an innovative strategy called GPT-Assisted Annotation. 
% where GPT-4 is used to automate the initial annotation process. The annotated data then underwent a meticulous review and manual verification conducted by two medical professionals. 
This approach harnessed the power of GPT-4 to automate the initial annotation process. Subsequently, the annotated data underwent a meticulous review and manual verification conducted by two medical professionals. Figure~\ref{fig:sample} shows an example question from CMExam and the annotation process. 

Furthermore, we benchmark the performance of general domain LLMs and medical domain LLMs on answer prediction (multiple-choice) and answer reasoning (open-ended) tasks of CMExam. This comprehensive assessment aims to highlight the strengths and weaknesses of various approaches in Chinese medical QA, with a focus on LLMs. 
The main findings of this benchmark are as follows:
\begin{itemize}[labelindent=\parindent,leftmargin=10pt, itemsep=0pt, parsep=0pt]
    \item GPT-4~\citep{OpenAI2023GPT4TR} demonstrates impressive zero-shot performance on the answer prediction task compared to other models, though still significantly lagging behind human performance.
    \item GPT-3.5~\citep{brown2020language} and GPT-4 generated reasonable answers on the answer reasoning task despite low BLEU and ROUGE scores. This is because they tended to generate short answers with reasonable quality.
    \item Existing medical domain LLMs, such as Huatuo~\citep{li2023huatuo} and DoctorGLM~\citep{Xiong2023DoctorGLMFY}, exhibit poor zero-shot performance on both tasks, indicating their limited coverage of medical knowledge and substantial room for improvement.
    \item Lightweight LLMs (e.g., ChatGLM~\citep{du2022glm}) fine-tuned on CMExam with supervision chieve performance close to GPT-3.5 on the answer prediction task. They also significantly outperform GPT-3.5 and GPT-4 on the reasoning task while having only ~3\% of the parameters of GPT-3.5.
\end{itemize}

% \begin{itemize}
%     \item GPT-4~\citep{OpenAI2023GPT4TR} demonstrates impressive zero-shot performance on the answer prediction task compared to other models, though still significantly lagging behind human performance.
%     \item GPT-3.5~\citep{brown2020language} and GPT-4 generated reasonable answers on the answer reasoning task despite low BLEU and ROUGE scores. They tend to generate short answers with reasonable quality, which may have led to their low BLEU scores.
%     \item Existing medical domain LLMs, such as Huatuo~\citep{li2023huatuo} and DoctorGLM~\citep{Xiong2023DoctorGLMFY}, exhibit poor zero-shot performance on both tasks, indicating their limited coverage of medical knowledge and substantial room for improvement.
%     \item Lightweight LLMs (e.g., ChatGLM~\citep{du2022glm}) fine-tuned on CMExam with supervision chieve performance close to GPT-3.5 on the answer prediction task. They also significantly outperform GPT-3.5 and GPT-4 on the reasoning task while having only ~3\% of the parameters of GPT-3.5.
%     \end{itemize} 

In summary, this study provides valuable insights into the performance of LLMs in medical contexts from multiple perspectives, benefiting both the artificial intelligence research community and the medical research community. Our findings contribute to a deeper understanding of the capabilities and limitations of LLMs in the medical domain. Additionally, the CMExam dataset and benchmark introduced in this study serve as valuable resources to inspire researchers to explore more effective ways of integrating medical knowledge into LLMs, ultimately enhancing their performance in medical applications.

\begin{table}[ht]
\centering
\caption{A review of medical QA datasets. $^*$ indicates availability of additional annotations with authoritative references, $^\dagger$ indicates availability of benchmarks, and $^\ddag$ indicates datasets with more than 50K questions}
\resizebox{\textwidth}{!}{%
\begin{tabular}{llcc}
\hline
\multirow{2}{*}{\textbf{Language}} & \multirow{2}{*}{\textbf{Data Source Type}} & \multicolumn{2}{c}{\textbf{Question Type}} \\ \cmidrule{3-4}
& & \textbf{Multiple Choice} & \textbf{Open-ended}\\ \hline
\multirow{ 2}{*}[-2em]{English} & Consumer Questions & \makecell[l]{MedMCQA  \citep{pal2022medmcqa} } &  \makecell[l]{LiveQA-Med  \citep{abacha2017trec} \\ CliCR$^\ddag$  \citep{vsuster2018clicr}\\HealthQA \citep{zhu2019healthqa}\\MEDIQA \citep{abacha2019mediqa}\\emrQA$^\ddag$ 
 \citep{pampari2018emrqa}\\MedQuaD \citep{ben2019medquad}\\
 MedicationQA$^*$ \citep{abacha2019medicationqa}\\
 MEDIQA-AnS \citep{savery2020MEDIQA-AnS}\\
 MASH-QA \citep{zhu2020MASH-QA}}\\\cline{2-4}
& Research, Books, or Exams & \makecell[l]{MEDQA$^\ddag$\citep{jin2021medqa}\\MMLU$^{\dagger\ddag}$ \citep{hendrycks2020mmlu}\\MedMCQA \citep{pal2022medmcqa}\\MultiMedQA$^{*\dagger}$ \citep{singhal2022multimedqa}}& \makecell[l]{ BioASQ  \citep{krithara2023bioasq}\\MultiMedQA$^{*\dagger}$ \citep{singhal2022multimedqa}} \\\hline
\multirow{ 2}{*}[-1em]{Chinese} & Consumer Questions & - & \makecell[l]{webMedQA$^{*\ddag}$ \citep{he2019webmedqa}\\cMedQA-v1.0$^\ddag$   \citep{zhang2017cmed-qa-1}\\ cMedQA-v2.0$^\ddag$   \citep{zhang2018cmed-qa-v2} \\ ChiMed  \citep{tian2019chimed} \\ Huatuo-26M$^{\dagger\ddag}$   \citep{li2023huatuo}} \\\cline{2-4}
& Research, Books, or Exams & \makecell[l]{MLEC-QA$^\ddag$   \citep{li2021mlec-qa} \\\textbf{CMExam}$^{*\dagger\ddag}$(ours)}& \makecell[l]{MLEC-QA$^\ddag$  \citep{li2021mlec-qa} \\\textbf{CMExam}$^{*\dagger\ddag}$(ours)} \\\hline
\end{tabular}%
}
\label{tab:dataset_review}
\end{table}

\section{Related Work}

\myparagraph{Medical Question-Answering Datasets}
% This section focuses on the comprehensiveness and objectiveness of medical QA datasets, as these factors heavily influence the reliability of these datasets in assessing large language models. 
Table \ref{tab:dataset_review} presents a summary of medical QA datasets published after 2017. In particular, we focus on categorizing the data source and question types of the different datasets. Most existing medical QA datasets adopt an open-ended format, primarily because they were constructed directly from consumer questions and answers from doctors. However, multiple-choice and fill-in-the-blank questions provide a more standardized and objective evaluation, and only a small portion of medical QA datasets have adopted these formats. Notable examples include CliCR \citep{vsuster2018clicr}, MEDQA \citep{jin2021medqa}, MMLU \citep{hendrycks2020mmlu}, MLEC-QA \citep{li2021mlec-qa}, and MedMCQA \citep{pal2022medmcqa}. Note that the multiple-choice questions in MultiMedQA \citep{singhal2022multimedqa} come from MEDQA, MedMCQA, and MMLU.

Data source types generally determine the reliability of a dataset. Consumer questions collected from web sources require human review to ensure the correctness of the answers. As datasets grow in size, quality control becomes increasingly challenging \citep{li2023huatuo}. In contrast, datasets built from case reports (e.g., CliCR), research literature (e.g., BioAsq \citep{krithara2023bioasq}), medical books, exams, and related practices (e.g., MMLU and MedMCQA) are often more reliable.

From Table \ref{tab:dataset_review}, we observe that there are few datasets based on multiple-choice questions from authoritative sources. This characteristic distinguishes CMExam from the MLEC-QA dataset, which is also derived from the Chinese National Medical Licensing Examination. In essence, CMExam has been meticulously crafted as a foundational benchmark dataset. It introduces question explanations for reasoning ability inspection, incorporates expansive annotation facets with authoritative references, and includes question-wise medical competencies and difficulty ratings calculated from human performance. These features make CMExam an indispensable resource for authoritative LLM performance assessment and meaningful human-machine comparisons. Table \ref{tab:unique_value_of_annotation} presents a list of innovations and characteristics of CMExam, which are discussed in detail in the following sections.

\myparagraph{Other Benchmark Datasets of Large Language Models}
The assessment of LLMs has witnessed significant progress, with the introduction of diverse benchmarks that evaluate different dimensions across multiple languages and models. Many datasets focus on assessing natural language understanding and reasoning capabilities of LLMs. RACE \citep{lai2017race} includes English exams for Chinese middle and high school students. TriviaQA \citep{joshi2017triviaqa} consists of question-answer pairs authored by trivia enthusiasts. DROP \citep{dua2019drop} evaluates reading comprehension with discrete reasoning and arithmetic components. GLUE \citep{wang2018glue} encompasses four existing NLU tasks, while SuperGLUE \citep{wang2019superglue} extends it with a more challenging benchmark of eight language understanding tasks. Other datasets, such as HellaSwag \citep{zellers2019hellaswag} and WinoGrande \citep{sakaguchi2021winogrande}, focus on commonsense reasoning. TruthfulQA \citep{lin2021truthfulqa} includes health, law, finance, and politics, to assess LLMs' ability to mimic human falsehoods, while MMCU \citep{zeng2023mmmcu} covers medical, legal, psychology, and education to evaluate multitask Chinese understanding. In addition to language understanding and reasoning, several datasets focus on specific subjects and topics, such as Python coding tasks \citep{chen2021Codex} and middle school mathematics questions \citep{cobbe2021GSM8K}. Notably, both C-Eval \citep{huang2023c} and M3KE \citep{liu2023m3ke} serve as multi-level multi-subject evaluation benchmarks, making them particularly suitable for assessing the capabilities of LLMs across multiple domains.

\section{The CMExam Dataset}

\label{dataset}
\myparagraph{Data Collection and Pre-processing}
% The Chinese National Medical Licensing Examination (CNMLE), also known as the Physician Qualification Examination, is a standardized exam that evaluates whether applicants for medical practitioner qualifications have the necessary professional knowledge and skills required for medical work. The examination is divided into two levels, namely, licensed physician and assistant licensed physician; each level is further divided into four categories, including clinical medicine, traditional Chinese medicine, stomatology, and public health. The CNMLE is comprised of two parts: a written test and a clinical skills assessment. The written component consists of multiple-choice questions covering an extensive range of medical subjects that include anatomy, physiology, pathology, pharmacology, clinical medicine, etc. The clinical skills assessment evaluates the applicant's ability to diagnose and treat patients in a simulated clinical environment.
CMExam comprises authentic past licensed physician exams in the Chinese National Medical Licensing Examination (CNMLE) collected from the Internet. The CNMLE, also known as the Physician Qualification Examination, is a standardized exam that assesses applicants' medical knowledge and skills in China. It includes a written test with multiple-choice questions covering various medical subjects and a clinical skills assessment simulating patient diagnosis and treatment.
We excluded questions that rely on non-textual information, including questions with external information such as images and tables, and questions with keywords "graph" and "table". Duplicate questions were removed from the dataset. In total, 96,161 questions, 68,119 of which were retained after pre-processing. The dataset was then randomly split into training/development/test sets with a ratio of 8:1:1. Each question in the dataset is associated with an ID, five candidate answers, and a correct answer. 85.24\% of questions have brief solution explanations and questions in the test set contain additional annotations.

\myparagraph{Data Annotation}
CMExam provides a comprehensive analysis of LLM performance through five additional annotation dimensions. The first dimension involves disease groups based on the 11th revision of the International Classification of Diseases (ICD-11) \citep{icd11}. ICD-11 is a globally recognized standard classification system for documenting and categorizing health conditions, consisting of 27 major disease groups. The second dimension comprises 36 clinical departments derived from the Directory of Medical Institution Diagnostic and Therapeutic Categories (DMIDTC) \footnote{~\url{http://www.nhc.gov.cn/fzs/s3576/201808/345269bd570b47e7aef9a60f5d17db97.shtml}}, published by the National Health Commission of China. DMIDTC is an authoritative guide used for categorizing and naming diagnostic and therapeutic subjects within healthcare institutes. In cases where the question cannot be successfully classified by ICD-11 or DMIDTC, the annotation is marked as "N/A". The third dimension refers to medical disciplines, which are categorized based on the List of Graduate Education Disciplinary Majors (2022) published by the Ministry of Education of the People's Republic of China\footnote[3]{~\url{http://www.moe.gov.cn/srcsite/A22/moe_833/202209/t20220914_660828.html}}. This dimension encompasses seven categories representing study majors used in universities. The fourth dimension was created by two medical professionals within the team to assess the primary medical competency tested by each associated question. It consists of four categories. The fifth dimension represents five potential difficulty levels of each question, determined by analyzing the correctness rate observed in human performance data collected alongside the questions. For detailed information on these additional annotations including their potential values, please refer to Table \ref{tab:icd11}, \ref{tab:clinical_dept}, \ref{tab:disciplines}, \ref{tab:competency}. And our proposed GPT-Assisted Annotation strategy is shown in supplementary materials. %Appendix \ref{details_of_additional_annotations}.

\begin{table}[t]
    \caption{Additional annotations of CMExam. }
    \centering
    \resizebox{1\textwidth}{!}{
    \begin{tabular}{llc}
    \hline
        Annotation Content & References & Unique values \\ 
    \hline
        Disease Groups & The 11th revision of ICD-11 & 27 \\ 
        Clinical Departments & The Directory of Medical Institution Diagnostic and Therapeutic Categories (DMIDTC) & 36 \\ 
        Medical Disciplines & List of Graduate Education Disciplinary Majors (2022) & 7 \\ 
        Medical Competencies & Medical Professionals & 4 \\ 
        Difficulty Level & Human Performance & 5 \\ 
    \hline
    \end{tabular}}
\label{tab:unique_value_of_annotation}
\end{table}

\myparagraph{Dataset Characteristics}
The CMExam dataset has several advantages over previous medical QA datasets regarding: 1)\textit{Reliability and Authenticity}: CMExam is sourced exclusively from the CNMLE that undergoes rigorous review and validation processes, ensuring its accuracy and adherence to established medical standards. 2) \textit{Standardization and Comprehensiveness}: CMExam includes both multiple-choice questions that ensure fair and objective evaluations of models' performance and question-wise open-ended reasoning that allows in-depth analysis and assessment of model reasoning abilities and comprehension. Despite the inherent absence of explanations within the CNMLE, we cross-referenced exam questions with solutions offered by diverse online medical examination preparation platforms, effectively enhancing the dataset's informational depth. CMExam reflects the comprehensive coverage of medical knowledge and reasoning required in clinical practice, as it is sourced from carefully designed national medical exams. The inclusion of five additional annotation dimensions enhances the dataset's rigor and offers valuable insights for in-depth evaluation and analysis. 3) \textit{Scale}: CMExam consists of over 60K high-quality questions, providing a large and reliable dataset.

% \footnote{~\url{https://www.yikaobang.com.cn/}} \footnote{~\url{http://www.jinyingjie.com/}} \footnote{~\url{https://www.lanjiyin.com.cn/}} 

\begin{figure}[ht]
  \centering
  \includegraphics[width=1\textwidth]{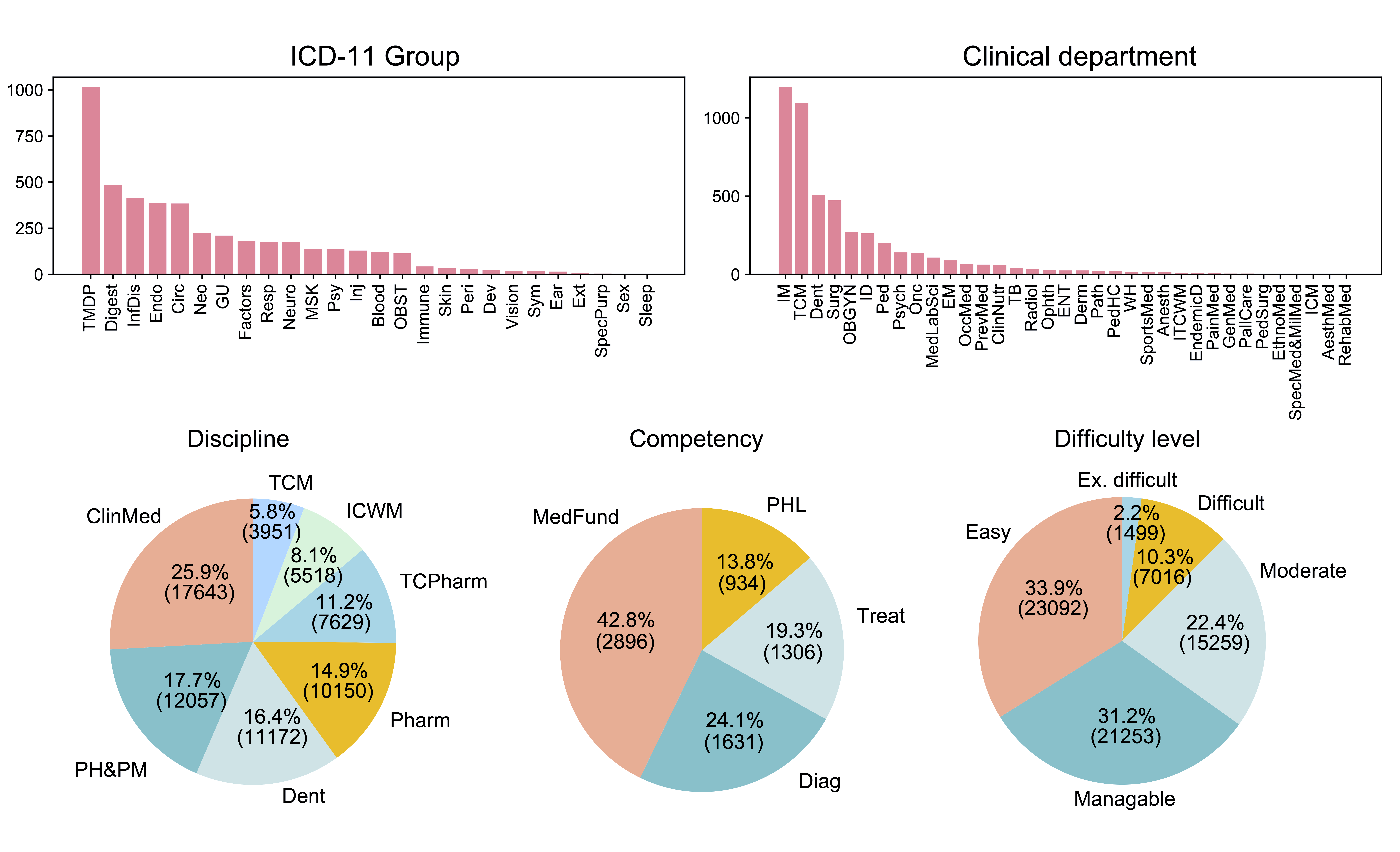}
  \caption{Additional CMExam statistics. For the question length distribution subplot, only the portion within IQR is shown.}
  \label{fig:distribution}
\end{figure}

\myparagraph{Data Statistics}
The dataset has a total of 68,119 questions, with 65,950 answers being single-choice and 2,169 being multiple-choice, with a maximum of five answer choices. Among all questions, 85.24\% have associated solution explanations \footnote{~\url{https://www.yikaobang.com.cn/},~\url{http://www.jinyingjie.com/},~\url{https://www.lanjiyin.com.cn/}}. Figure \ref{fig:distribution} shows additional statistics visualization and more basic statistics of CMExam can be seen in supplementary materials.
Within the test set, 4,493 questions (65.97\%) have corresponding disease group annotations. The most prevalent disease group is Traditional Medicine Disease Patterns (TMDP), followed by Digestive System Diseases, Certain Infectious (Digest) and Parasitic Diseases (InfDis), Endocrine, Nutritional, or Metabolic Diseases (Endo), and Circulatory System Diseases (Circ).
For the associated clinical department annotations, 4,965 questions (72.90\%) have been assigned values. The two most frequently represented clinical departments are Internal Medicine (IM) and Traditional Chinese Medicine (TCM), with Dentistry (Dent) and Surgery (Surg) following closely.
Every question in the test set has been labeled with a discipline, where Clinical Medicine (ClinMed) comprises the largest proportion. Additionally, each question has been categorized into a competency area, with Medical Fundamentals (MedFund) being the predominant category.
The difficulty levels of the questions align with common exam patterns, with a greater number of easy questions and a smaller number of hard questions.

\section{Benchmarks}

\subsection{Baselines, Settings, and Metrics}
\myparagraph{Model Selection}
The LLMs we benchmarked on the CMExam can be divided into two groups based on domains:
% encapsulates a diverse range of QA strategies, and can be systematically grouped into two categories: 
1) \textit{General Domain LLMs}: This group comprises GPT3.5/4~\citep{brown2020language, OpenAI2023GPT4TR}, ChatGLM~\citep{du2022glm,zeng2023glm-130b}, LLaMA~\citep{Touvron2023LLaMAOA}, Alpaca~\citep{alpaca}, and Vicuna~\citep{vicuna2023}. These models are general-purpose language models trained on a massive amount of general-purpose corpora;
2) \textit{Medical Domain LLMs}: 
 This group can be further divided into two subgroups. The first subgroup consists of representative LLMs specifically designed for the medical domain, including DoctorGLM~\citep{Xiong2023DoctorGLMFY} and Huatuo~\citep{wang2023huatuo}. DoctorGLM is a healthcare-specific language model initialized with ChatGLM-6B parameters and further fine-tuned on Chinese medical dialogues extracted from ChatGPT. Huatuo, on the other hand, is a knowledge-enhanced model, which builds upon the LLaMA architecture and is additionally supervised-fine-tuned with knowledge-based instruction data harvested from the Chinese medical knowledge graph (CMeKG).
 The second subgroup comprises medical LLMs that were constructed through supervised fine-tuning of LLMs using the CMExam training set. This subgroup includes models fine-tuned on BERT~\citep{Devlin2019BERTPO}, RoBERTa~\citep{roberta}, PromptCLUE~\citep{clueai2023promptclue} (T5-based), BART ~\citep{shao2021cpt}, Huatuo, ChatGLM, LLaMA, Alpaca, and Vicuna.
% \subsubsection{Traditional Open-domain Multiple-choice QA}

% \subsubsection{General Domain Model}

% \subsubsection{Domain Specific Model}

\myparagraph{Human Performance}
To effectively gauge the medical proficiency of LLMs, incorporating a measure of human performance into the benchmarking process is of paramount importance.  Therefore, during data collection, we preserved the accuracy of human responses for each question. Human performance is estimated by computing a weighted average of response accuracy within each dimension, with weights determined by the number of respondents. This design ensures a robust comparison of LLMs' performance relative to human capabilities, particularly when larger respondent samples contribute to a question's accuracy. 

\myparagraph{Experimental Setting}
\label{Experimental Setting}
% In our experimental setting, we utilized various models for evaluation. 
For GPT models, we leveraged OPENAI's API to access the GPT-3.5-turbo and GPT-4-0314 models, given that their open-source variants are currently unavailable. The LLaMA, Alpaca, and Vicuna models were used in their respective 7B versions, while ChatGLM was evaluated using its publicly accessible 6B version. Additionally, we performed fine-tuning on open-source models using the CMExam dataset. We used P-tuning V2 \citep{Liu2021PTuningVP} for ChatGLM-6B, with the length of prefix tokens set to 128, and the learning rate set to 2e-2, LoRA \citep{Hu2021LoRALA} for LLaMA, Alpaca, Vicuna, and Huatuo models, with the rank set to 8, alpha set to 16, and dropout at 0.05. 
% Fine-tuning involved specific hyperparameters such as batch size, maximum input and target length, and number of epochs. 
For BERT models, we followed the fine-tuning methods outlined in \citep{Devlin2019BERTPO}, with batch size set to 16, learning rate set to 2e-4, hidden dropout probability set to 0.4, and maximum input length set to 192. The fine-tuning processes for all models except BERT involved a batch size of 64, a maximum input length, and a target length of 256. All fine-tuning was performed using NVIDIA V100 GPUs for 10 epochs.

% In relation to GPT models, we leveraged OPENAI's API to access the GPT-3.5-turbo and GPT-4-0314 models, given that their open-source variants are currently unavailable. For the LLaMA, Alpaca, and Vicuna models, we employ their respective 7B versions, taking advantage of the public availability of these models. Regarding ChatGLM, we adopt its publicly accessible 6B version for our evaluations.
% Additionally, we fine-tuned several open-source models on the proposed CMExam dataset for evaluation. P-tuning V2 \citep{Liu2021PTuningVP} was utilized to perform fine-tuning on the ChatGLM-6B model, with the length of prefix tokens set to 128 and a learning rate of 2e-2. LoRA \citep{Hu2021LoRALA} was used to fine-tune the LLaMA, Alpaca, Vicuna, and Huatuo models, with rank set to 8, alpha set to 16, and dropout at 0.05. We use the Huatuo-LLaMA-med model in our experiments. Other hyper-parameters used in the fine-tuning processes included a batch size of 64, a maximum input length and target length of 256, and a total of 10 epochs. All evaluations and fine-tuning were performed using NVIDIA V100 GPUs. The fine-tuning approaches for the BERT models followed the methods outlined in \citep{Devlin2019BERTPO} for training the reader models. The BERT model's fine-tuning batch size is set to 16, with the learning rate set to 2e-4 and the hidden dropout probability set to 0.4. The highest-scoring contexts for each question are retrieved from the retriever. 

\myparagraph{Metrics}
We assess model performance on multiple choice questions using accuracy and weighted F1 score. These metrics are commonly employed in information retrieval and question-answering tasks to evaluate model performance. For the open-ended solution explanations of CMExam, BLEU \citep{Papineni2002BleuAM} and ROUGE \citep{Lin2003AutomaticEO} were used to evaluate the discrepancy between model-generated explanations and ground truth. 
% These are well-regarded measures in NLP, offering a quantitative evaluation of the agreement between generated and reference texts.

\subsection{Results and Analysis}

\myparagraph{Overall Comparison}
% 在实验中，我们针对答案预测和答案解析两个任务上，测评了传统方法、通用大模型和在医疗场景finetuing的大模型在CMExam数据集上的性能。如图所示。针对答案选择任务，可以看到相比传统算法，通用领域的大模型能够达到更好的效果，并且随着模型参数的增大，效果有明显提升。比如ChatGLM-6B能够达到26%的准确率，175B的GPT-3.5达到47%的准确率。另外，经过使用医学数据finetune的模型，效果也能有较为客观的提升，例如在CMExam数据集上Finetune的ChatGLM-FT，使用6B的模型参数，就可以达到46%的准确率，接近175B和更大参数规模的GPT-4的性能。另外，可以看到人类的准确率能够达到71.6%，这说明在CMExam数据集上，大模型与人类之间还存在着较大的差距。另外，针对答案解释任务，可以看到GPT系列模型在BLEU指标的上性能较差，这是因为他们生成的解释一般较短，而BLEU指标会对较短文本长度进行惩罚。但是可以看到他们在ROUGE指标上更占优势。通过finetune，LLM能够生成更为合理的解释，比如CHatGLM-FT在BLEU-1和BLEU-4上指标能够达到30.9和18.85，在ROUGE指标上能够达到44.0，31.5和29.5.
We first assessed the performance of general domain LLMs and medical domain LLMs for answer prediction and reasoning tasks. The results are displayed in Table \ref{tab:overall_comparison}.
For the answer prediction task, GPT-4 significantly outperforms other methods, demonstrating a zero-shot performance with an accuracy of 61.6\% and an F1 score of 0.617. While a performance gap still exists when compared to human performance (which stands at 71.6\% accuracy), it's noteworthy that this gap has been greatly reduced from what was observed with GPT-3.5.
Among lightweight, general domain LLMs, ChatGLM outperforms LLaMA, Alpaca, and Vicuna, likely attributable to their limited coverage of the Chinese corpus. This restriction seemingly hampers their ability to provide accurate responses to CMExam queries. Furthermore, a noticeable deficiency in zero-shot performance is evident in lightweight medical domain LLMs such as Huatuo, owing to their restricted medical corpus diversity, which hampers the acquisition of broad medical knowledge and accurate interpretation of CMExam questions. Our findings suggest that finetuning models with CMExam enhance their performance. For instance, with an accuracy of 45.3\%, ChatGLM-CMExam is comparable to GPT-3.5's performance, despite utilizing only about 3\% of the parameters employed by GPT-3.5.
It is noteworthy that encoder-only LLMs, such as BERT and RoBERTa, remain a robust baseline for answer prediction tasks. Their performance can par with, or even exceed, that of certain decoder-only LLMs, such as LLaMA-CMExam and Alpaca-CMExam, despite having fewer parameters. 
 % indicate that general domain models outperform traditional algorithms in the answer selection task, and their performance significantly improves as the model parameters increase. For instance, 6B ChatGLM achieved an accuracy of 26.4\%, while 175B GPT-3.5 attained an accuracy of 47.0\%. In addition, we found that finetuning models with medical data enhanced their performance. For example, ChatGLM-CMExam, which was finetuned on the CMExam dataset using 6B model parameters, achieved an accuracy rate of 46.3\%, comparable to the performance of GPT-4, which utilizes substantially more parameters. The LLaMA and Alpaca models have inadequate coverage for Chinese, which hinders their ability to provide answers to certain questions, leading to an unsatisfactory performance on CMExam.      

For the solution explanation task, we observe that GPT models performed poorly on the BLEU metric, likely due to their tendency to generate short explanations. However, they exhibited an advantage on the ROUGE metric. As DoctorGLM is unable to return answer options according to the prompt, we only report its performance in the solution explanation task. Through finetuning, LLM was able to generate more reasonable explanations. For instance, ChatGLM-CMExam achieved scores of 31.10 and 18.94 on BLEU-1 and BLEU-4, respectively, and scores of 43.94, 31.48, and 29.39 on the ROUGE metrics.

% \usepackage{multirow}
% \usepackage{graphicx}
\begin{table}[]
\centering
\caption{Overall comparison on CMExam dataset. We \textbf{bold} the best result and \underline{underline} the second best result.}
\resizebox{\textwidth}{!}{%
\begin{tabular}{llcccccccc}
\toprule
\multirow{2}{*}{Model type} & \multirow{2}{*}{Models} & \multirow{2}{*}{size} & \multicolumn{2}{c}{Prediction} & \multicolumn{5}{c}{Reasoning } \\
\cmidrule(lr){4-5} \cmidrule(lr){6-10}
 &  &  & Acc (\%) & F1 (\%) & BLEU-1 & BLEU-4 & ROUGE-1 & ROUGE-2 & ROUGE-L \\
\midrule
\multirow{6}{*}{General Domain} 
 & GPT-3.5-turbo & 175B &\underline{46.4±0.6} & \underline{46.1±0.7} & 3.56±0.67 & 1.49±0.51 & 33.80±0.19 & 16.39±0.18 & 14.83±0.13 \\
 & GPT-4 & - &  \textbf{61.6±0.1} & \textbf{61.7±0.1} & 0.17±0.00 & 0.06±0.00 & 29.74±0.09 & 14.84±0.04 & 11.51±0.03 \\
 & ChatGLM & 6B & 26.3±0.0 & 25.7±0.1 & 16.51±0.08 & 5.00±0.06 & 35.18±0.11 & 15.73±0.05 & 17.09±0.13 \\
 & LLaMA & 7B & 0.4±0.0 & 0.3±0.0 & 11.99±0.03 & 5.70±0.0 & 27.33±0.06 & 11.88±0.03 & 10.78±0.04 \\
 & Vicuna & 7B & 5.0±0.0 & 4.8±0.1 & 20.15±0.01 & 9.26±0.01 & 38.43±0.02 & 16.90±0.01 & 16.33±0.01\\
 & Alpaca & 7B & 8.5±0.0 & 8.4±0.0 & 4.75±0.00 & 2.50±0.00 & 22.52±0.00 & 9.54±0.00 & 8.40±0.00 \\
\midrule
\multirow{10}{*}{Medical Domain}
& Huatuo & 7B & 12.9±0.0 & 7.0±0.0 & 0.21±0.00 & 0.12±0.00 & 25.11±0.08 & 11.56±0.04 & 9.73±0.02 \\
 & MedAlpaca & 7B  & 20.0±0.0 & 10.7±0.0 &0.00±0.00  &0.00±0.00  &1.90±0.00  & 0.04±0.00  & 0.52±0.03  \\
% & Huatuo-LLaMA-Literature & 7B & 2.1 &0.035 & 2.73 & 0.79 & 17.92 & 4.56 & 7.84 \\
% & Huatuo-Alpaca-med & 7B & 1.3 & 0.018 & 17.48 & 9.53 & 39.42 & 18.75 & 16.70 \\
% & Huatuo-Alpaca-Literature & 7B & 1.2 & 0.020 & 12.87 & 5.81 & 28.66 & 12.12 & 11.97 \\
 & DoctorGLM & 6B & -  & - & 9.43±0.09  & 2.65±0.03 & 21.11±0.03 & 6.86±0.01 & 9.99±0.06 \\
 \cmidrule(lr){2-10}

% & PromptCLUE-base-CMExam & 0.1B  & - & - &18.75±0.08  &6.65±0.05  &40.88±0.11  &21.90±0.11  &18.31±0.11  \\
% & Bart-base-chinese-CMExam & 0.1B  & - & - &23.00±0.40  &10.35±0.16  &44.33±0.09  &24.29±0.09  &20.80±0.09  \\
% & Bart-large-chinese-CMExam & 0.1B  & - & - &26.37±0.18  &11.65±0.08  &44.92±0.12  &24.34±0.12  &21.75±0.03  \\

& PromptCLUE-base-CMExam & 0.1B  & - & - &18.75±0.08  &6.65±0.05  &40.88±0.11  &21.90±0.11  &18.31±0.11 \\
& Bart-base-chinese-CMExam & 0.1B  & - & - &23.00±0.40 &10.35±0.16  &44.33±0.09 &24.29±0.09  &20.80±0.09 \\
& Bart-large-chinese-CMExam & 0.1B  & - & - &26.37±0.18  &11.65±0.08  &44.92±0.12  &24.34±0.12  &21.75±0.03  \\
  & BERT-CMExam & 0.1B  & 31.8±0.2 & 31.2±0.2 &-  &-  &-  &-  &-  \\
 % & Chinese-bert-wwm-ext-CMExam & 0.1B & 24.1 & 0.240 &-  &-  &-  &- &-  \\
 & RoBERTa-CMExam & 0.3B & 37.1±0.1 & 36.7±0.4 &-  &-  &- &- &-  \\
 % & Huatuo-Alpaca-CMExam & 7B & 23.4 & 0.267  & 30.49 & 18.08 & 46.85 & 27.96 & 24.26 \\∂
 & MedAlpaca-CMExam & 7B  & 30.5±0.1 & 30.4±0.1 &16.35±0.80  &9.78±0.47  &44.31±0.85  &\underline{27.05±0.50}  &\underline{24.55±0.43}  \\
 & Huatuo-CMExam & 7B & 28.6±0.5 & 29.3±0.2  & 29.04±0.01 & 16.72±0.03 & 43.85±0.24 & 25.36±0.22 & 21.72±0.24 \\
 & ChatGLM-CMExam & 6B & 45.3±1.4 & 45.2±1.4 & \textbf{31.10±0.23} & \textbf{18.94±0.12} & 43.94±0.28 & \textbf{31.48±0.14} & \textbf{29.39±0.14} \\
 & LLaMA-CMExam & 7B & 18.3±0.5 & 20.6±0.5 & 29.25±0.23 & 16.46±0.10 & \textbf{45.88±0.04} & 26.57±0.04 & 23.31±0.02 \\
 & Alpaca-CMExam & 7B & 21.1±0.6 & 24.9±0.4 & 29.57±0.10 & 16.40±0.12 & \underline{45.48±0.12} & 25.53±0.18 & 22.97±0.06 \\
 & Vicuna-CMExam & 7B & 27.3±0.5 & 28.2±0.3 & \underline{29.82±0.03} & \underline{17.30±0.01} & 44.98±0.16 & 26.25±0.13 & 22.44±0.09 \\
 \midrule
Random & Random & - & 3.1±0.2 & 5.1±0.3 & - & - & - & - & - \\
\midrule
Human Performance & Human volunteers & - & 71.6 & - & - & - & - & - & - \\

\bottomrule
\end{tabular}%
}
\label{tab:overall_comparison}
\end{table}

\myparagraph{Results by Disease Groups}
Drawing upon ICD-11 annotations (26 categories), we conducted an analysis of the performance of several LLMs across various categories. To mitigate the potential impact of random variability resulting from the number of questions, we limited our analysis to categories containing more than 100 questions. According to Table \ref{tab:comparing_disease}, LLMs have uneven performance and significant gaps in knowledge. GPT-4’s accuracy ranges from 74.4\% for \textit{Neo} to 44.3\% for \textit{TCMDP}, GPT-3.5's accuracy ranges from 63.9\% for \textit{Neo} to 31.0\% for \textit{TCMDP} and ChatGLM-CMExam's accuracy ranges from 54.7\% for \textit{Psy} to 42.9\% for \textit{RESP}.

\begin{table}[t]
    % \begin{minipage}{0.48\linewidth}
    \centering
    \caption{Comparing disease classifications.}
    \resizebox{0.8\textwidth}{!}{%
    \begin{tabular}{cccccc}
    \toprule
    \textbf{Categories} & \textbf{GPT-4} & \textbf{GPT-3.5} & \textbf{ChatGLM} & \textbf{ChatGLM-CMExam} & \textbf{Average}   \\ 
    \midrule
    Neo & 74.4±2.2 & 63.9±1.4 & 32.4±1.6 & 51.9±0.2 & 55.6±0.8 \\
    Psy & 74.0±0.7 & 62.0±1.7 & 33.3±1.3 & 54.7±0.8 & 56.0±0.9 \\
    Factors & 70.0±1.0 & 57.5±1.4 & 28.0±1.1 & 51.1±1.4 & 51.6±0.5 \\
    MSK & 65.9±0.8 & 53.8±0.8 & 29.2±0.4 & 53.5±0.0 & 50.6±0.4 \\
    GU & 69.2±0.4 & 52.1±1.1 & 30.0±0.2 & 49.5±0.9 & 50.2±0.3 \\
    Inj & 65.9±2.3 & 45.7±1.3 & 37.2±2.9 & 49.1±1.8 & 49.5±1.4 \\
    Circ & 68.8±0.3 & 49.3±0.7 & 30.9±0.7 & 47.0±0.3 & 49.0±0.2 \\
    Endo & 70.6±1.1 & 49.4±1.1 & 25.5±0.8 & 46.1±0.4 & 47.9±0.2 \\
    Digest & 67.0±1.0 & 48.8±1.4 & 26.2±0.7 & 49.4±1.1 & 47.8±0.4 \\
    InfDis & 66.0±0.5 & 49.2±0.8 & 27.5±0.6 & 48.2±0.8 & 47.7±0.4 \\
    Neuro & 64.4±1.2 & 48.7±3.1 & 28.6±0.4 & 45.3±1.3 & 46.7±1.1 \\
    OBST & 63.5±0.3 & 45.0±2.4 & 25.7±0.9 & 49.4±0.3 & 45.9±0.5 \\
    BLOOD & 69.4±0.3 & 45.3±1.4 & 18.9±1.6 & 43.3±0.7 & 44.2±0.4 \\
    Resp & 62.7±0.8 & 44.3±1.4 & 24.5±0.3 & 42.9±0.0 & 43.6±0.7 \\
    N/A & 60.0±0.1 & 46.8±0.3 & 24.9±0.2 & 42.5±0.1 & 43.5±0.1 \\
    TCMDP & 44.3±0.9 & 31.0±0.6 & 24.2±0.4 & 47.9±0.0 & 36.9±0.6 \\
    \bottomrule
    \end{tabular}%
    }
    \label{tab:comparing_disease}
    % \end{minipage}
    % \begin{minipage}{0.48\linewidth}
    \centering
    \caption{Comparing clinical department.}
    \resizebox{0.8\textwidth}{!}{%
    \begin{tabular}{cccccc}
    \toprule
    \textbf{Categories} & \textbf{GPT-4} & \textbf{GPT-3.5} & \textbf{ChatGLM} & \textbf{ChatGLM-CMExam}  & \textbf{Average} \\ 
    \midrule
    EM & 67.4±0.2 & 49.8±0.7 & 36.3±0.4 & 50.2±0.5 & 50.9±0.1 \\
    OBGYN & 66.4±1.0 & 51.7±1.5 & 28.6±0.5 & 52.0±0.0 & 49.7±0.3 \\
    IM & 70.2±0.6 & 51.8±0.8 & 26.0±1.1 & 47.9±0.9 & 49.0±1.0 \\
    ID & 67.4±1.9 & 49.5±3.3 & 26.1±1.9 & 49.6±3.8 & 48.2±1.2 \\
    Surg & 63.6±0.8 & 49.5±1.5 & 28.8±0.5 & 47.7±0.9 & 47.4±1.5 \\
    ClinNutr & 68.3±2.4 & 48.3±2.9 & 23.9±1.1 & 47.8±0.5 & 47.1±0.7 \\
    MedLabSci & 69.2±0.6 & 48.3±2.0 & 29.0±1.5 & 40.8±0.6 & 46.8±0.2 \\
    Ped & 64.5±0.0 & 47.2±1.4 & 26.7±2.1 & 41.9±5.5 & 45.1±1.7 \\
    N/A & 62.6±0.2 & 48.6±1.1 & 24.6±0.4 & 44.3±0.9 & 45.0±1.0 \\
    Ophth & 60.9±0.5 & 39.1±0.8 & 21.8±0.8 & 54.0±0.2 & 44.0±0.8 \\
    OccMed & 61.5±4.3 & 38.5±1.6 & 31.3±4.3 & 41.5±3.3 & 43.2±2.5 \\
    DENT & 54.9±2.0 & 41.2±1.6 & 27.9±0.8 & 43.5±0.9 & 41.9±1.0 \\
    TCM & 43.1±1.3 & 31.4±1.3 & 24.5±1.9 & 45.8±4.4 & 36.2±0.6 \\
    ENT & 41.3±0.8 & 28.0±0.6 & 29.3±0.1 & 26.7±0.1 & 31.3±0.5 \\
    ICM & 33.3±0.0 & 11.1±15.7 & 0.0±0.0 & 11.1±15.7 & 13.9±4.8 \\
    \bottomrule
    \end{tabular}%
    }
    \label{tab:comparing_specialty_and_department}
    % \end{minipage}
\end{table}

\myparagraph{Results by Clinical Departments}
To compare model performance regarding the clinical department dimension (36 categories), we only analyzed categories with more than 50 questions to ensure result representativeness. Results presented in Table \ref{tab:comparing_specialty_and_department} highlight that the models show relatively high accuracy on questions associated with commonly encountered departments, such as Emergency Medicine (\textit{EM}), Internal Medicine (\textit{IM}) and Surgery (\textit{Surg}). Their accuracy on questions associated with rarer departments, such as Traditional Chinese Medicine (\textit{TCM}). There is a marked discrepancy in the average accuracy among different departments, with the highest being 50.9\% and the lowest being only 13.9\%. This observation suggests there are notable variations in medical knowledge and reasoning approaches among different departments. Consequently, it may be necessary to examine specific optimization strategies for different departments.

% To evaluate the performance of the LLMs across various disciplines, we conducted an analysis of seven fields, including clinical medicine (ClinMed), dentistry (Dent), integrated Chinese and Western medicine (ICWM), public health and preventive medicine (PH\&PM), pharmacy (Pharm), traditional Chinese medicine (TCM), and traditional Chinese Pharmacy (TCPharm). As illustrated in Tab.\ref{tab:comparing_medical_discipline}, the results showed that the performances of LLMs in \textit{TCM}, \textit{TCPharm}, and \textit{Pharm} were the poorest, with all accuracy rates below 40\%. This suggests that the LLMs may have had the least exposure to these types of data. On the other hand, \textit{ClinMed} and \textit{Ph\&PM} had the highest accuracy rates, likely due to the availability of more relevant data. Overall, these performance differences among different disciplines highlight the unique data features and challenges in each field, calling for targeted model optimization and improvement.
\myparagraph{Results by Medical Disciplines}
Then, we evaluated LLM performance across seven medical disciplines. As depicted in Table \ref{tab:comparing_medical_discipline}, the performance of LLMs across disciplines such as Traditional Chinese Medicine (\textit{TCM}), Traditional Chinese Pharmacy (\textit{TCPharm}), and Pharmacy (\textit{Pharm}) was notably subpar, with all accuracy rates falling below 42\%. This pattern suggests a potential deficiency in the exposure of these models to data within these categories. Conversely, disciplines such as \textit{ClinMed} and \textit{Ph\&PM} demonstrated higher accuracy rates, likely due to the abundance of relevant data. The observed variability in performance across different disciplines underscores the distinctiveness of data characteristics and complexities inherent to each field, thereby advocating for discipline-specific model optimizations and enhancements.

\begin{table}[t]
% \begin{minipage}{0.48\linewidth}
\centering
\caption{Comparing medical discipline.}
\resizebox{0.8\textwidth}{!}{%
\begin{tabular}{cccccc}
\toprule
\textbf{Categories} & \textbf{GPT-4} & \textbf{GPT-3.5} & \textbf{ChatGLM} & \textbf{ChatGLM-CMExam} & \textbf{Average} \\ 
\midrule
    ClinMed & 67.9±0.1 & 51.4±0.4 & 27.3±0.3 & 48.9±0.4 & 48.8±0.7 \\
    PH\&PM & 68.2±0.4 & 52.7±1.7 & 26.2±0.3 & 47.3±1.0 & 48.6±0.5 \\
    ICWM & 56.1±0.1 & 40.0±2.3 & 29.4±0.8 & 53.6±0.7 & 44.8±0.9 \\
    Dent & 59.5±0.7 & 43.9±1.9 & 28.5±1.1 & 45.3±0.6 & 44.3±0.3 \\
    Pharm & 61.1±0.4 & 46.3±0.5 & 23.2±0.2 & 37.0±0.1 & 41.9±0.3 \\
    TCM & 53.5±0.4 & 35.9±0.2 & 24.1±0.3 & 49.1±0.0 & 40.6±1.1 \\
    TCPharm & 45.4±1.2 & 35.6±0.1 & 24.1±1.0 & 43.1±0.4 & 37.1±0.5 \\
\bottomrule
\end{tabular}%
}
\label{tab:comparing_medical_discipline}
% \end{minipage}
% \begin{minipage}{0.48\linewidth}
\centering
\caption{Comparing LLMs' competencies.}
\resizebox{0.8\textwidth}{!}{%
\begin{tabular}{cccccc}
\toprule
\textbf{Categories} & \textbf{GPT-4} & \textbf{GPT-3.5} & \textbf{ChatGLM} & \textbf{ChatGLM-CMExam} & \textbf{Average} \\ 
\midrule
Diag & 70.1±5.5 & 50.9±2.1 & 30.9±2.8 & 51.6±1.0 & 50.9±1.4 \\
PHL & 64.2±0.7 & 50.0±0.5 & 26.8±0.3 & 49.6±0.1 & 47.6±0.3 \\
Treat & 56.5±0.5 & 43.0±1.1 & 25.7±0.2 & 47.4±0.6 & 43.2±0.8 \\
MeFund & 58.3±0.3 & 44.6±0.7 & 23.9±0.5 & 41.6±0.4 & 42.1±0.9 \\
N/A & 54.8±0.2 & 30.4±0.4 & 23.7±0.1 & 38.5±0.2 & 36.9±0.3 \\
\bottomrule
\end{tabular}%
}
\label{tab:comparing_competency}
% \end{minipage}
\end{table}

% This comprehensive evaluation aimed to provide a holistic assessment of the models' capabilities in addressing the complex challenges and requirements of diverse healthcare contexts. As shown in Tab.\ref{tab:comparing_competency}, the average accuracy of the LLMs in the domain of mastery of fundamental knowledge (\textit{MedFund}) is the lowest, with an average score of only 38.7\%. This finding suggests that the models, which were trained on general scenarios, had insufficient exposure to medical data. Although fine-tuning the models led to some improvement in accuracy, the limited number of samples used in fine-tuning implies that the models still require more medical scenario data to enhance their performance. Nonetheless, it is noteworthy that the average accuracy of predicting public health laws and ethics (\textit{PHL}) common sense questions is relatively high, with a notable achievement of 43.8\%. Furthermore, the LLMs demonstrate their capability on accurate disease diagnosis (\textit{Diag}) with an accuracy of 46.3\%, suggesting that their reasoning abilities are consistent with the anticipated performance.
\myparagraph{Results by Competencies} Evaluations based on medical competency areas aimed at a higher-level understanding of model capability in solving medical problems. As indicated in Table \ref{tab:comparing_competency}, the lowest average accuracy across LLMs was observed within the domain of mastering Medical Fundamentals (\textit{MedFund}), with a meager average score of 42.1\%. This result demonstrates that these models, predominantly trained on general textual data, have inadequate exposure to medical-specific data. While fine-tuning did provide some improvement, these models could benefit from additional medical scenario data to further augment their performance. It is worth highlighting that the average accuracy in the domain of Public Health Laws and Ethics (\textit{PHL}) was reasonably high, notably achieving an average of 47.6\%. In addition, the LLMs showcased their proficiency in accurate disease diagnosis.

% During the experiment, we assessed the performance of the LLMs in tackling questions of varying levels of difficulty. The results depicted in Tab.\ref{tab:comparing_difficulty_of_the_questions} reveal a clear trend where the accuracies of the model diminish with an increase in question complexity. This observation indicates that more intricate questions require a broader knowledge base and complex reasoning, which presents a challenge to the LLMs, mirroring human performance.
\begin{table}
\centering
\caption{Results by question difficulty.}
\resizebox{0.8\textwidth}{!}{%
\begin{tabular}{cccccc}
\toprule
\textbf{Categories} & \textbf{GPT-4} & \textbf{GPT-3.5} & \textbf{ChatGLM} & \textbf{ChatGLM-CMExam} & \textbf{Average} \\ 
\midrule
Easy & 74.6±0.1 & 58.5±0.6 & 31.4±0.2 & 61.5±0.3 & 56.5±0.4 \\
Manageable & 63.9±0.2 & 47.4±0.7 & 25.9±0.5 & 46.1±0.3 & 45.8±0.6 \\
Moderate & 51.3±0.6 & 36.8±0.8 & 23.0±0.4 & 34.5±0.6 & 36.4±0.7 \\
Difficult & 36.4±0.9 & 26.2±0.7 & 18.9±0.5 & 24.3±0.9 & 26.5±0.6 \\
Extremely difficult & 27.2±1.0 & 21.4±2.2 & 15.8±1.0 & 12.2±1.1 & 19.1±1.1 \\
\bottomrule
\end{tabular}%
}
\label{tab:comparing_difficulty_of_the_questions}
\end{table}

\myparagraph{Results by Question Difficulty}
To evaluate model performance in tackling questions of varying levels of difficulty, we conducted experiments regarding the question difficulty dimension, which was calculated based on human exam-taker performance. As shown in Table \ref{tab:comparing_difficulty_of_the_questions}, there's an evident trend where model accuracies decrease as question complexity rises. This pattern suggests that more sophisticated questions demand an extensive knowledge base and complex reasoning, which are challenging for the LLMs, thus reflecting patterns observed in human performance.

% \begin{table}[t]

% As depicted in Fig.\ref{fig:comparing_length_of_question}, it is evident that the LLMs exhibit higher accuracy when solving problems of lengths ranging between 60 and 90. Conversely, their performance tends to decline when tackling problems that are either too short or too long. Furthermore, we observe the impacts of question length vary for different LLMs. The GPT models' performance gradually improves as the length of the problem increases, with the optimal range being between 50 and 90. On the other hand, ChatGLM-CMExam exhibits significant fluctuations in performance under different lengths and tends to underperform in comparison to the GPT models when dealing with longer problems.
\begin{wrapfigure}{r}{0cm}
  \centering
  \includegraphics[width=0.4\textwidth]{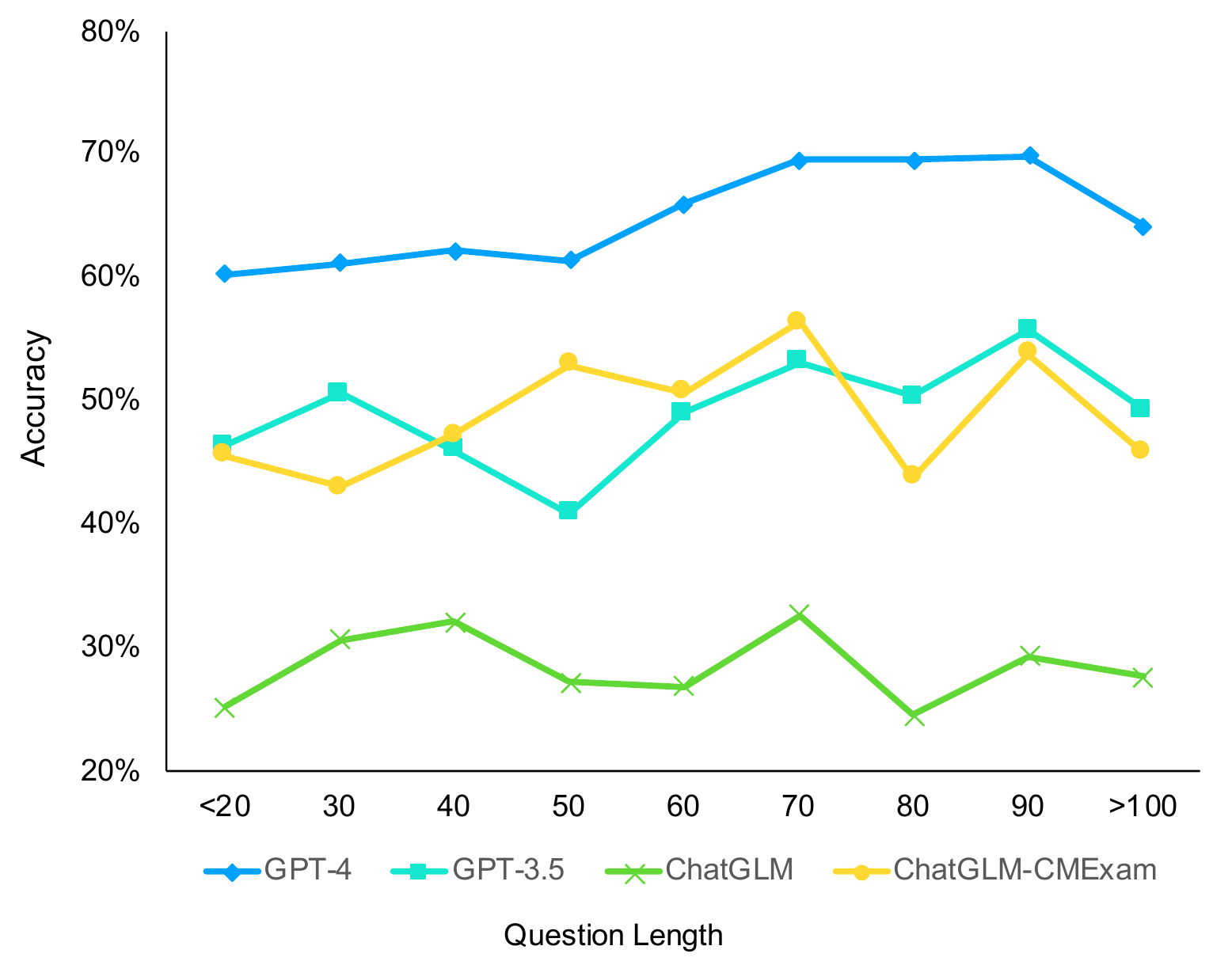}
  \caption{Results stratified by 
  question length.}
  \label{fig:comparing_length_of_question}
\end{wrapfigure}

\myparagraph{Results by Question Length}
Finally, to investigate if model performance is associated with input lengths, we compared their performance regarding question lengths. Figure \ref{fig:comparing_length_of_question} illustrates that Large Language Models (LLMs) generally show higher accuracy with problem lengths between 60 and 90. However, their performance seems to falter with problems that are either too short or overly long. Additionally, we noticed that the effect of question length varies across different LLMs. For instance, GPT models tend to incrementally improve as the problem length expands, performing optimally within the 50 to 90 range. Conversely, ChatGLM-CMExam's performance fluctuates noticeably with varying lengths, and it tends to fall short compared to GPT models when addressing longer problems.

% \begin{figure}[t]

% \subsubsection{Case Studies}

% \section{Ethics and Limitations}

% \begin{figure}[htbp]
%     \begin{minipage}{0.6\textwidth}
%         \centering
%         \captionsetup{type=table} % 设置标题类型为表格
%         \caption{Results by question difficulty.}
%         % 这里插入你的表格代码
%         \resizebox{1.0\textwidth}{!}{%
%         \begin{tabular}{cccccc}
%         \toprule
%         \textbf{Categories} & \textbf{GPT-4} & \textbf{GPT-3.5} & \textbf{ChatGLM} & \textbf{ChatGLM-CMExam} & \textbf{Average} \\ 
%         \midrule
%         Easy & 74.7\% & 59.3\% & 31.2\% & 61.3\% & 56.6\% \\
%         Manageable & 63.8\% & 48.3\% & 26.1\% & 46.3\% & 46.1\% \\
%         Moderate & 50.6\% & 37.3\% & 23.1\% & 34.9\% & 36.5\% \\
%         Difficult & 37.5\% & 25.4\% & 18.3\% & 24.9\% & 26.5\% \\
%         Ex. difficult & 27.5\% & 19.8\% & 16.8\% & 11.5\% & 18.9\% \\
%         \bottomrule
%         \end{tabular}}%
%         \label{tab:comparing_difficulty_of_the_questions}
%     \end{minipage}%
%     \begin{minipage}{0.4\textwidth}
%         \centering
%         \includegraphics[width=\textwidth]{Comparing_question_length.pdf}
%         \caption{Results stratified by 
%   question length.}
%         \label{fig:comparing_length_of_question}
%     \end{minipage}
% \end{figure}

\section{Conclusion and Discussions}

% Future Work}
In this work, we developed CMExam, a dataset sourced from the stringent Chinese National Medical Licensing Examination, featuring 60,000+ multiple-choice questions, with detailed explanations. CMExam ensures reliability, validity, and adherence to medical standards. It also demonstrates the practicality of employing GPT-4 to automate the annotation process, which strikes a harmonious balance between efficiency and cost-effectiveness while maintaining the desired level of accuracy and reliability of the annotation. Utilizing this large and reliable corpus, we tested several LLMs for answer selection and reasoning tasks. A performance gap was observed between LLMs and human experts, signaling the need for additional LLM research. CMExam's standardization and comprehensiveness also ensure objective evaluations of models while enabling in-depth analysis of their reasoning capabilities. The questions cover a wide spectrum of medical knowledge, augmented with five additional annotation dimensions for rigorous evaluation. This study aims to spur further exploration of LLMs in medicine by providing a comprehensive benchmark for their evaluation. We anticipate CMExam to contribute significantly to future advancements of LLMs, particularly in handling medical question-answering tasks.

\myparagraph{Limitations}
\label{limitations}
% We recognize several limitations in our study.
Firstly, while CMExam is derived from meticulously designed medical examinations, our process of excluding questions requiring non-textual information may inadvertently affect the balance of the remaining questions, potentially introducing unexpected biases. It is critical to acknowledge this aspect while interpreting any findings or analyses conducted using this dataset. 
% Additionally, while our dataset is larger in scale compared to most existing medical datasets, it is still relatively small when compared to general QA datasets like SuperGLUE \citep{wang2019superglue}. To address this limitation, techniques such as data augmentation or knowledge distillation from publicly-available pre-trained medical models, such as PubMedBERT \citep{pubmedbert} and ClinicalBERT \citep{huang2019clinicalbert}, can potentially be employed in the future. When expanding and constructing medical datasets, it is crucial to emphasize the importance of close collaboration with clinical experts to ensure real-world impact. 
Furthermore, the current BLEU and ROUGE metrics primarily evaluate the explanation task, but these measures are insufficient for assessing the reasonableness of the answer.
% since model performance is greatly influenced by the length of the generated answer. 
In future work, we will incorporate human evaluation to provide a more comprehensive assessment of the models.

\myparagraph{Ethics}
\label{ethics}
% \mysubparagraph{General Ethical Conduct}: 
CMExam is a dataset derived from the Chinese National Medical Licensing Examination, which aligns with numerous datasets containing similar National Medical Licensing Examinations \citep{li2021mlec-qa,hendrycks2020mmlu,jin2021medqa,pal2022medmcqa,singhal2022multimedqa}. We have ensured adherence to applicable legal and ethical guidelines during data collection and use. The authenticity and accuracy of the exam questions have been thoroughly verified, providing a reliable basis for evaluating LLMs. Please note that the CMExam dataset is intended for academic and research purposes only. Any commercial use or other misuse that deviates from this purpose is expressly prohibited. We urge all users to respect this stipulation in the interest of maintaining the integrity and ethical use of this valuable resource.

\myparagraph{Societal Impacts} While CMExam aims to enhance LLM evaluations in the medical field, it should not be misused for assessing individual medical competence or for patient diagnosis. Conclusions drawn from models trained on this dataset should acknowledge its limitations, especially given its single source and the specific context of the CNMLE. The use of this dataset should strictly be limited to research purposes to avoid potential misuse.

\bibliographystyle{ACM-Reference-Format}
\bibliography{main_and_supp}

%%%%%%%%%%%%%%%%%%%%%%%%%%%%%%%%%%%%%%%%%%%%%%%%%%%%%%%%%%%%
\appendix
\newpage
\section{Appendix}
\begin{CJK*}{UTF8}{gbsn}

\subsection{Abbreviations, Full Names, and Translations of Additional Annotations}
\label{details_of_additional_annotations}
This section presents four tables of additional annotations that contain translation. It showcases abbreviations, full English names, and Chinese names for each group in each annotation dimension. Table \ref{tab:icd11} showcases all disease groups included in the 11th revision of the International Classification of Diseases (ICD-11). We present the disease group in the same order found on the official website. Table \ref{tab:clinical_dept} offers a classification of 36 clinical departments derived from the Directory of Medical Institution Diagnostic and Therapeutic Categories. Table \ref{tab:disciplines} presents a breakdown of medical disciplines based on the List of Graduate Education Disciplinary Majors published by the Ministry of Education of the People's Republic of China. This categorization comprises seven study majors used in universities. Table \ref{tab:competency} provides all groups of areas of medical competency assessed in Chinese medical licensing exams.

\begin{table}[H]
\centering
\caption{ICD-11 Groups}
\resizebox{\textwidth}{!}{
\begin{tabular}{clll}
% \begin{tabular}{lp{6cm}p{4cm}}
\hline
Code& Abbreviation & Full English Name & Chinese Name \\
\hline
01& InfDis & Certain infectious or parasitic diseases & 某些感染性疾病或寄生虫病 \\
02& Neo & Neoplasms & 肿瘤 \\
03& Blood & Diseases of the blood or blood-forming organs & 血液或造血器官疾病 \\
04& Immune & Diseases of the immune system & 免疫系统疾病 \\
05& Endo & Endocrine, nutritional or metabolic diseases & 内分泌、营养或代谢疾病 \\
06& Psy & Mental, behavioural or neurodevelopmental disorders & 精神、行为或神经发育障碍 \\
07& Sleep & Sleep-wake disorders & 睡眠-觉醒障碍 \\
08& Neuro & Diseases of the nervous system & 神经系统疾病 \\
09& Vision & Diseases of the visual system & 视觉系统疾病 \\
10& Ear & Diseases of the ear or mastoid process & 耳或乳突疾病 \\
11&Circ & Diseases of the circulatory system & 循环系统疾病 \\
12&Resp & Diseases of the respiratory system & 呼吸系统疾病 \\
13 & Digest & Diseases of the digestive system & 消化系统疾病 \\
14 & Skin & Diseases of the skin & 皮肤疾病 \\
15& MSK & Diseases of the musculoskeletal system or connective tissue & 肌肉骨骼系统或结缔组织疾病 \\
16& GU & Diseases of the genitourinary system & 泌尿生殖系统疾病 \\
17& Sex & Conditions related to sexual health & 性健康相关情况 \\
18& OBST & Pregnancy, childbirth or the puerperium & 妊娠、分娩或产褥期 \\
19& Peri & Certain conditions originating in the perinatal period & 起源于围生期的某些情况 \\
20& Dev & Developmental anomalies & 发育异常 \\
21& Sym & Symptoms, signs or clinical findings, not elsewhere classified & 症状、体征或临床所见，不可归类在他处者 \\
22& Inj & Injury, poisoning or certain other consequences of external causes & 损伤、中毒或外因的某些其他后果 \\
23&Ext & External causes of morbidity or mortality & 疾病或死亡的外因 \\
24&Factors & Factors influencing health status or contact with health services & 影响健康状态或与 \\
25&SpecPurp & Codes for special purposes & 用于特殊目的的编码 \\
26&TCMDP & Supplementary Chapter Traditional Medicine Conditions - Module I & 补充章传统医学病证-模块1 \\
V&FuncAssess & Supplementary section for functioning assessment & 功能评定补充部分 \\
X&ExtCodes & Extension Codes & 扩展码 \\
-&N/A & Not Applicable & 不符合 \\
\hline
\end{tabular}}
\label{tab:icd11}
\end{table}

\begin{table}[H]
\centering
\caption{Medical Disciplines}
\resizebox{0.6\textwidth}{!}{
\begin{tabular}{lll}
\hline
Abbreviation & Full English Name & Chinese Name \\
\hline
ClinMed & Clinical Medicine & 临床医学 \\
Dent & Dentistry & 口腔医学 \\
ICWM & Integrated Chinese and Western Medicine & 中西医结合 \\
PH\&PM & Public Health and Preventive Medicine & 公卫预防 \\
Pharm & Pharmacy & 药学 \\
TCM & Traditional Chinese Medicine & 中医学 \\
TCPharm & Traditional Chinese Pharmacy & 中药学 \\
\hline
\end{tabular}}
\label{tab:disciplines}
\end{table}

\begin{table}[H]
\centering
\caption{Areas of competencies}
\resizebox{0.8\textwidth}{!}{
\begin{tabular}{lll}
\hline
Abbreviation & Full English Name & Chinese Name \\
\hline
Diag & Disease Diagnosis and Differential Diagnosis & 疾病诊断和鉴别诊断 \\
MedFund & Medical Fundamentals & 医学基础知识 \\
N/A & Not Applicable & 不符合 \\
PHL & Public Health Law and Ethics & 公共卫生法律伦理 \\
Treat & Disease Treatment & 疾病治疗 \\
\hline
\end{tabular}}
\label{tab:competency}
\end{table}

\begin{table}[H]
\centering
\caption{Clinical Departments}
\resizebox{1\textwidth}{!}{
\begin{tabular}{lll}
\hline
Abbreviation & Full English Name & Chinese Name \\
\hline
AesthMed & Aesthetic Medicine & 医疗美容科 \\
Anesth & Anesthesiology & 麻醉科 \\
ClinNutr & Clinical Nutrition & 临床营养科 \\
Dent & Dentistry & 口腔科 \\
Derm & Dermatology & 皮肤科 \\
EM & Emergency Medicine & 急诊医学科 \\
EndemicD & Endemic Disease & 地方病科 \\
ENT & Otolaryngology & 耳鼻咽喉科 \\
EthnoMed & Ethnic Medicine & 民族医学科 \\
GenMed & General Medicine & 全科医疗 \\
ICM & Intensive Care Medicine & 重症医学科 \\
ID & Infectious Diseases & 传染科 \\
IM & Internal Medicine & 内科 \\
ITCWM & Integrated Traditional Chinese and Western Medicine & 中西医结合科 \\
MedLabSci & Medical Laboratory Science & 医学检验科 \\
N/A & Not Applicable & 不符合 \\
OBGYN & Obstetrics and Gynecology & 妇产科 \\
OccMed & Occupational Medicine & 职业病科 \\
Onc & Oncology & 肿瘤科 \\
Ophth & Ophthalmology & 眼科 \\
PainMed & Pain Medicine & 疼痛科 \\
PallCare & Palliative Care & 临终关怀科 \\
Path & Pathology & 病理科 \\
Ped & Pediatrics & 儿科 \\
PedHC & Pediatric Health Care & 儿童保健科 \\
PedSurg & Pediatric Surgery & 儿童外科 \\
PrevMed & Preventive Medicine & 预防保健科 \\
Psych & Psychiatry & 精神科 \\
PT & Physical Therapy & 理疗科 \\
Radiol & Radiology & 医学影像科 \\
RehabMed & Rehabilitation Medicine & 康复医学科 \\
SpecMed\&MilMed & Special Medical and Military Medicine & 特种医学与军事医学科 \\
SportsMed & Sports Medicine & 运动医学科 \\
Surg & Surgery & 外科 \\
TB & Tuberculosis & 结核病科 \\
TCM & Traditional Chinese Medicine & 中医科 \\
WH & Women's Health & 妇女保健 \\
\hline
\end{tabular}}
\label{tab:clinical_dept}
\end{table}

\newpage
\subsection{Instructions for Pre-annotation}
\label{instruction_for_preannotation}
\begin{figure}[H]
  \centering
  \includegraphics[width=1\textwidth]{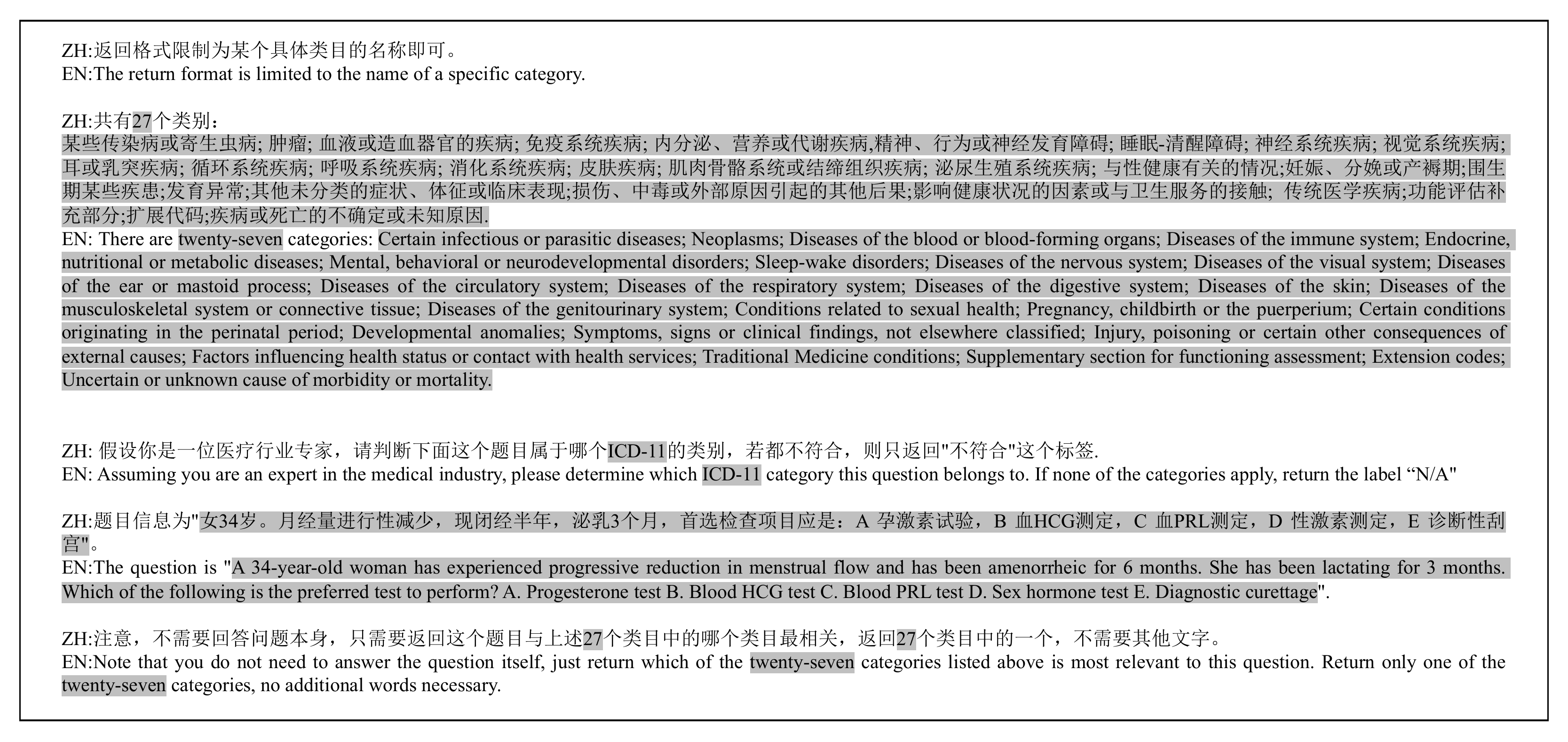}
  \caption{Pre-annotation Instructions for Disease Groups.}
  \label{fig:instruction_of_pre-annotation_disease_groups}
\end{figure}

\begin{figure}[H]
  \centering
  \includegraphics[width=1\textwidth]{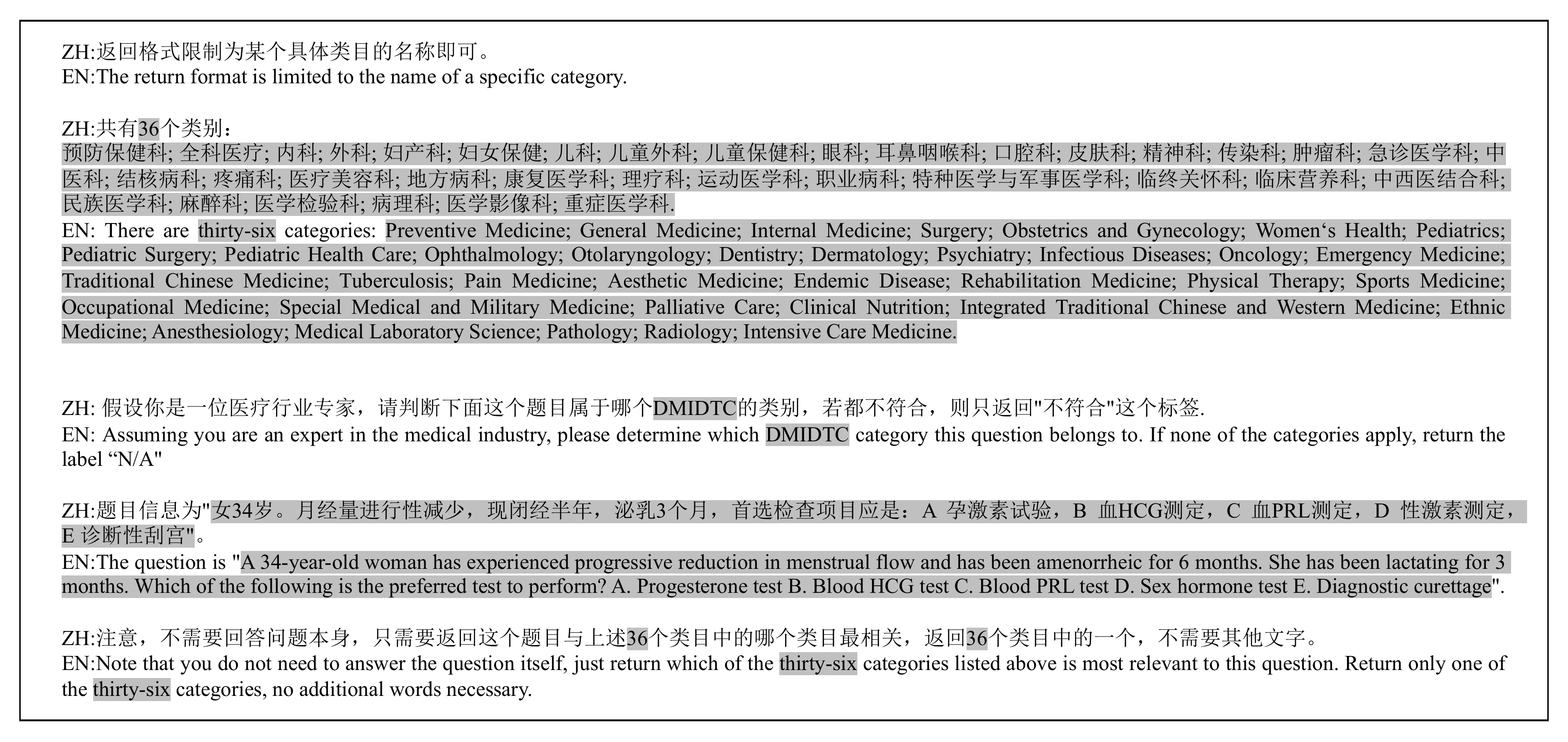}
  \caption{Pre-annotation Instructions for Clinical Departments.}
  \label{fig:instruction_of_pre-annotation_clinical_departments}
\end{figure}

\begin{figure}[H]
  \centering
  \includegraphics[width=1\textwidth]{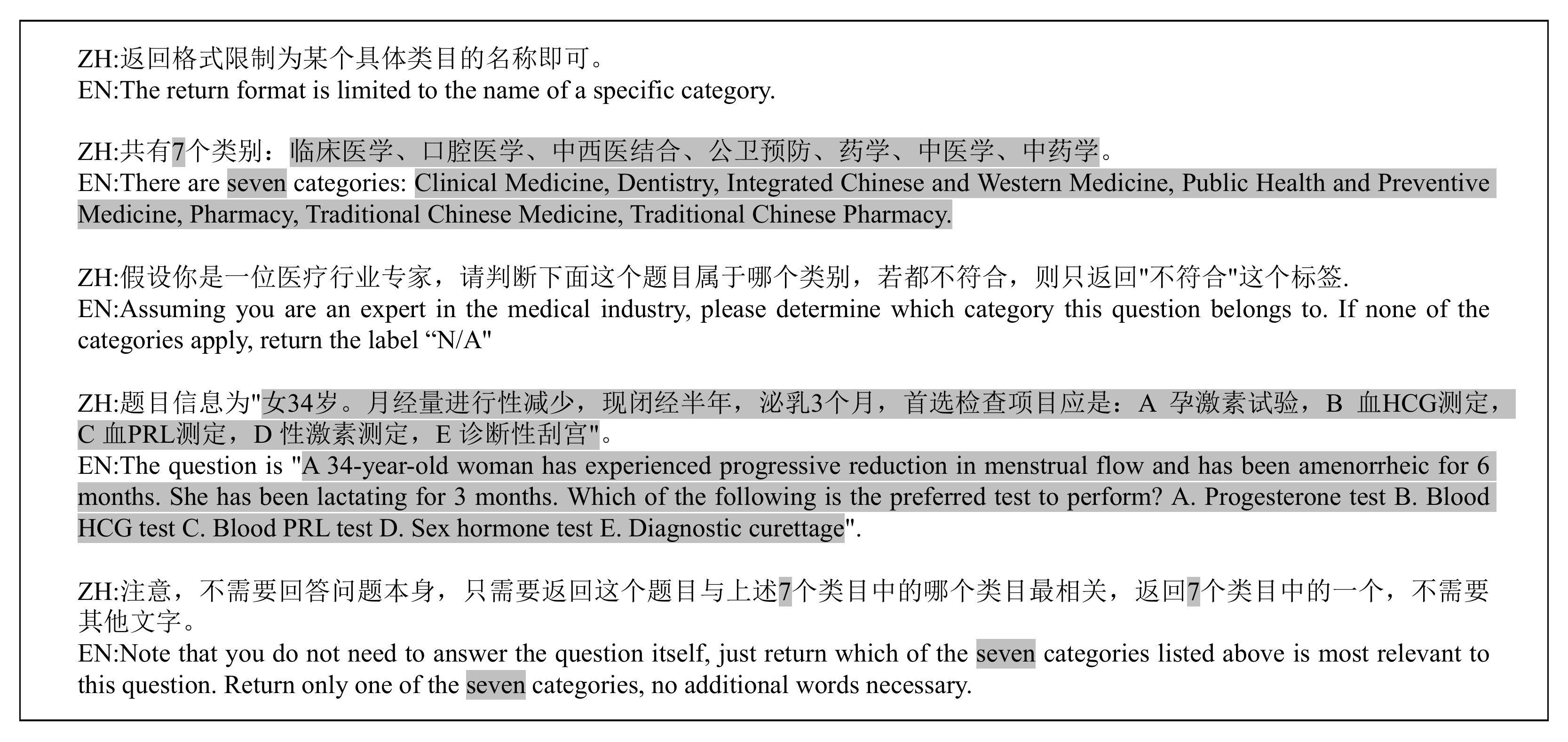}
  \caption{Pre-annotation Instructions for Medical Disciplines.}
  \label{fig:instruction_of_pre-annotation_medical_disciplines}
\end{figure}

\begin{figure}[H]
  \centering
  \includegraphics[width=1\textwidth]{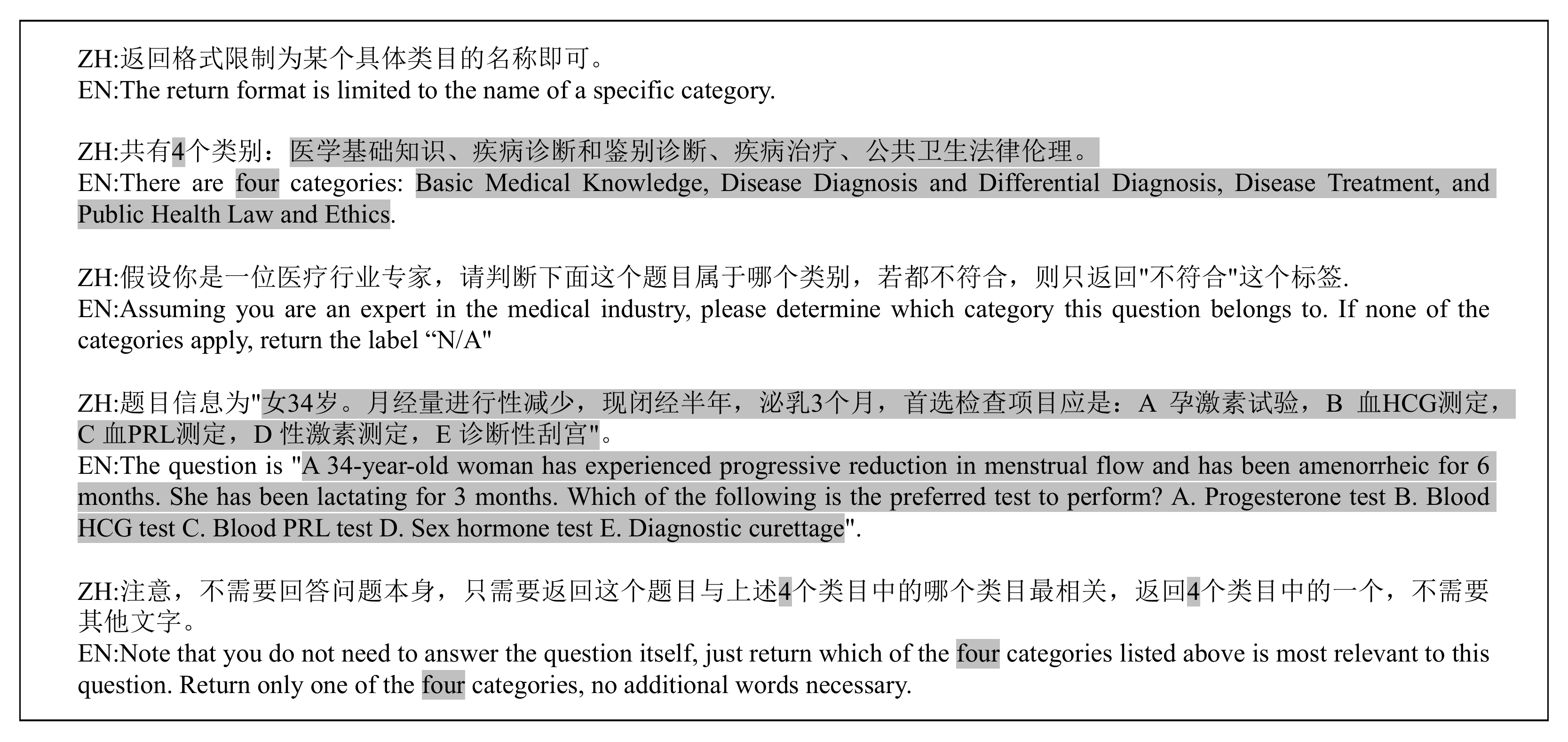}
  \caption{Pre-annotation Instructions for Areas of Competencies.}
  \label{fig:instruction_of_pre-annotation_areas_of_competencies}
\end{figure}

%在本小节，我们介绍了在实验中使用GPT4进行预标注时所使用的instructions。如图所示，首先我们对于GPT4的返回结果进行限制，要求只返回具体类目。然后针对本文中所涉及到的5个addtional annotation维度，在标注时分别填入每个维度下的所有类目信息。然后提供具体题目信息，最后再对GPT4的返回结果进行二次限制，从而提高预标注结果的有效性。在实际标注过程中，需要在灰色背景处填上具体的类目和题目信息。
In this section, we present instructions used to pre-annotate CMExam test set data using GPT4. As shown in Figure \ref{fig:instruction_of_pre-annotation_disease_groups},\ref{fig:instruction_of_pre-annotation_clinical_departments},\ref{fig:instruction_of_pre-annotation_medical_disciplines},\ref{fig:instruction_of_pre-annotation_areas_of_competencies}, we first constrained the output from GPT4 to return only specific categories. We then annotated each of the five additional annotation dimensions relevant to this study with all the category information for each dimension. Next, we provided specific prompt information and finally, we performed filtering on the GPT4 output to improve the effectiveness of pre-annotation. During the actual annotation process, specific categories and prompt information should be filled in the grey background areas.

\subsection{Analysis of Model Generation Ability}
%我们在图中展示了部分模型对于CMExam数据集中医疗问题的解释。可以看到，GPT-4和GPT-35会生成相对简短且合理的解释，这也解释了为什么他们在BLUE指标上不占优。而Vicuna, LLaMA,Huotuo等模型的重复生成现象较为明显，Alpaca只是将选项进行了重复，并没有给出任何解释。通过在CMExam数据集上进行finetune，可以看到重复生成的现象明显降低，并且解释的合理性也有明显提升，例如ChatGLM-CMExam模型会类似于solution explanation一样对每个选项进行分析。不过部分模型的解释仍然存在不合理性，例如LLaMa-CMExam, Aplaca-CMExam和Vicuna-CMExam，这可能是因为这些模型是用通用数据进行训练的，对于医疗专业领域的知识存在一定的不足，这也说明了训练一个医疗domain大模型的意义。

\begin{figure}[H]
  \centering
  \includegraphics[width=1.0\textwidth]
  {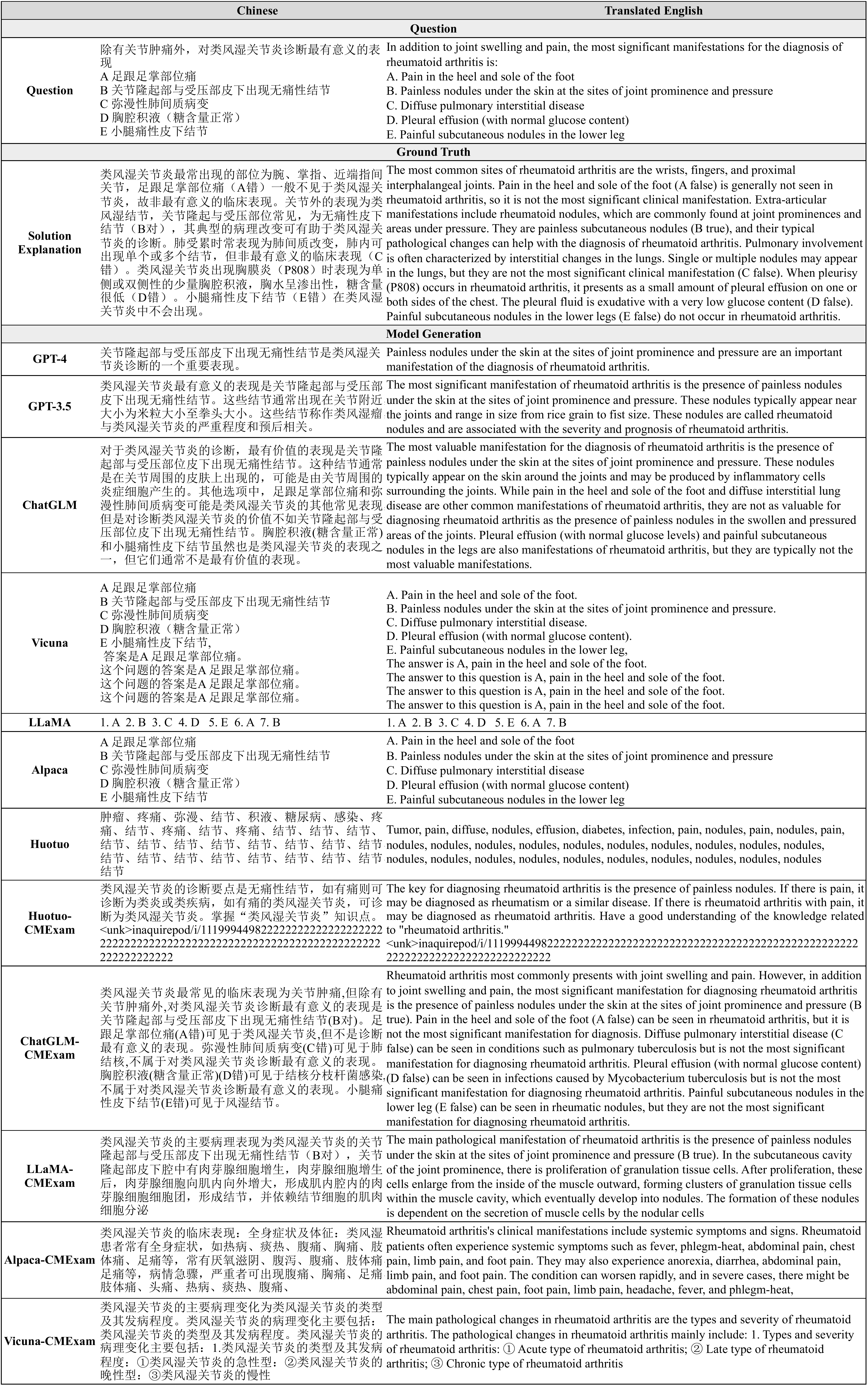}
  \caption{A case study of LLMs' generated explanations.}
  \label{fig:case_study}
\end{figure}

In Figure \ref{fig:case_study}, we present partial explanations generated by various models for a medical question from the CMExam dataset. Notably, GPT-4 and GPT-3.5 produce concise and sensible explanations, which may account for the lower BLUE scores. Conversely, models like Vicuna, LLaMA, and Huotuo demonstrate a more prominent repetition phenomenon, while Alpaca simply duplicates the provided options without providing an explanation.

Fine-tuning models on the CMExam dataset significantly reduces the repetition phenomenon and improves the overall reasonableness of the explanations. For instance, the ChatGLM-CMExam model analyzes each option in a similar manner to the solution explanation. However, some models still generate unreasonable explanations, as observed in LLaMA-CMExam, Alpaca-CMExam, and Vicuna-CMExam. This could be attributed to their training on generic data and lack of specific knowledge in the medical domain. This underscores the significance of training large language models with a focus on the medical domain.

% \subsection{Diagnostic Ability Analysis}

% \newpage
\subsection{Analysis of Model Generation Correctness}
% 我们随机选择几个模型答案预测正确的sample，邀请医学专家人工检查其生成的解释是否是正确的，而不仅仅是关注模型预测的结果正确性，最终发现虽然模型答案预测正确，但是还是存在一部分的sample，在进行答案解释时存在几种错误情况，专家主要将这些错误情况分为三类：解释与答案不相关、生成重复解释、解释不准确。统计如图所示，可以看到GPT models解释准确的sample数量均超过45，即占90%以上比例。而ChatGLM以及ChatGLM-CMExam会存在部分解释错误的情况，主要包含解释不准确以及不相关。我们在图中也对部分错误情况进行展示。

To assess the accuracy of model-generated explanations, we conducted a study using a randomly selected sample of 50 cases in which the Language Models (LLMs) correctly predicted the answers. Medical experts were then invited to manually verify the correctness of the explanations, focusing not only on the accuracy of the answer predictions but also on the quality of the accompanying explanations.

\begin{figure}[H]
  \centering
  \includegraphics[width=0.6\textwidth]{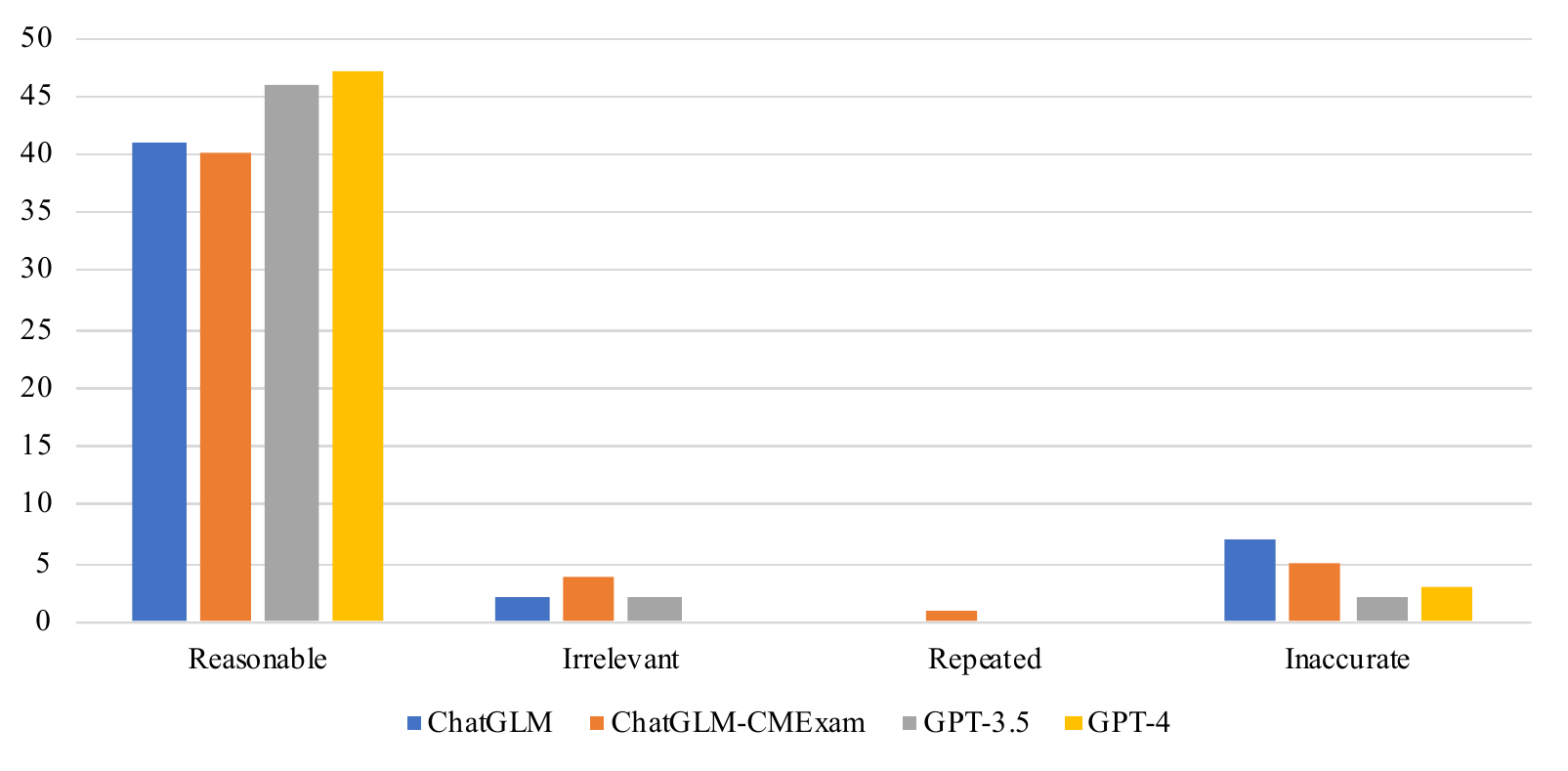}
  \caption{Correctness analysis.}
  \label{fig:correct_analysis}
\end{figure}

\begin{figure}
  \centering
  \rotatebox{270}{
    \begin{minipage}{\textheight}
      \includegraphics[width=1.0\linewidth]{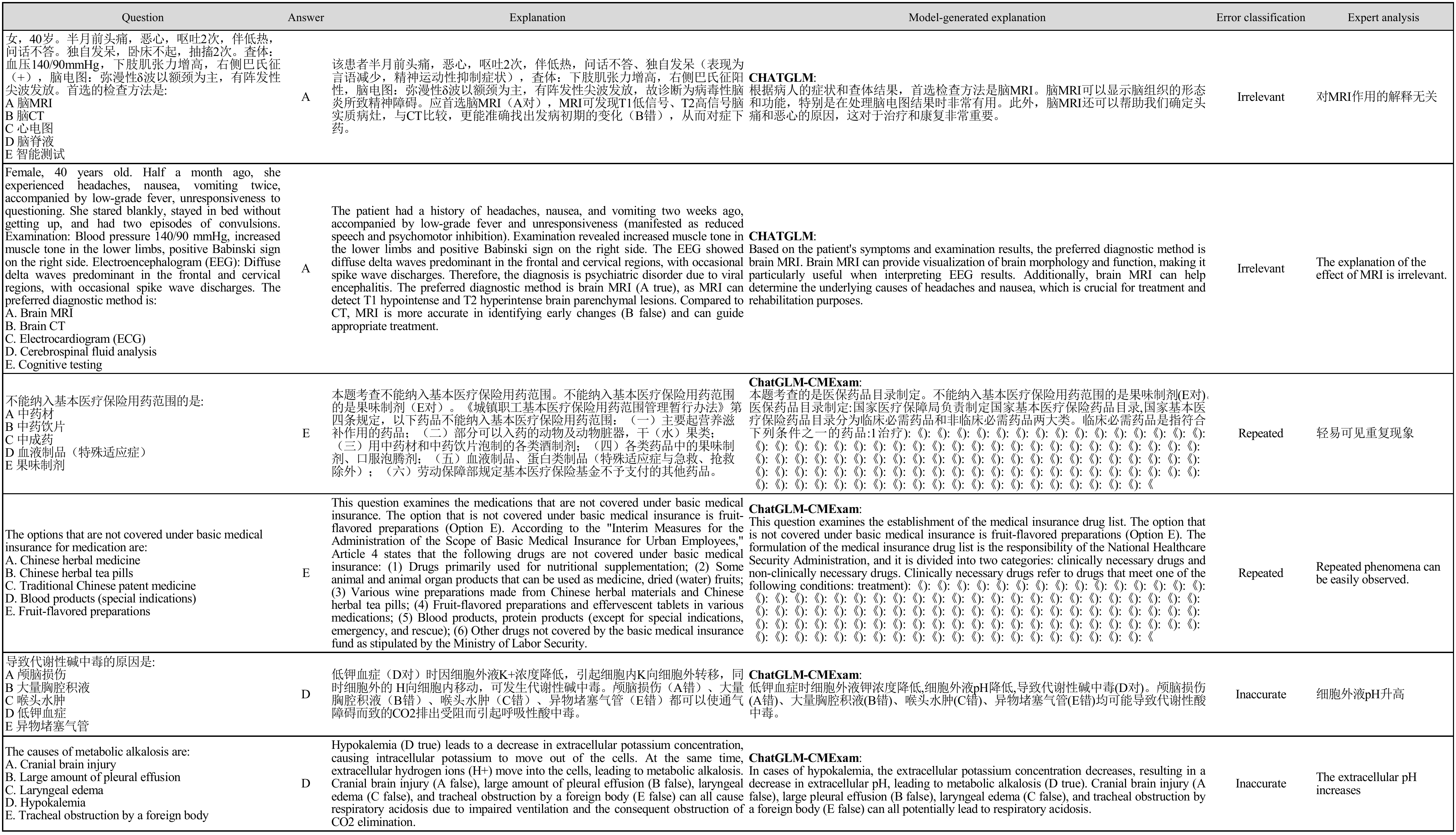}
      \caption{Examples of incorrect model-generated explanation.}
      \label{fig:correct_analysis_case}
    \end{minipage}
  }
\end{figure}

Our investigation revealed that despite the correct answer predictions by the models, certain samples exhibited errors in their corresponding explanations. These errors were categorized by the experts into three groups: explanations that were irrelevant, repeated, or inaccurate. The statistics presented in Figure \ref{fig:correct_analysis} demonstrate that the number of samples with accurate explanations generated by the GPT models exceeded 45, accounting for over 90\% of the total. However, it is important to note that both the ChatGLM and ChatGLM-CMExam models may produce some erroneous explanations, primarily consisting of inaccuracies and irrelevance. We have included examples of these incorrect explanations in Figure \ref{fig:correct_analysis_case}.
% \newpage
% \begin{figure}
%   \centering
%   \includegraphics[width=\linewidth]{correct_analysis.pdf}
%   \caption{第一部分}
%   \label{fig:part1}
% \end{figure}

% \begin{figure}
%   \ContinuedFloat
%   \centering
%   \includegraphics[width=\linewidth]{correct_analysis_case.pdf}
%   \caption{第二部分(Continued)}
%   \label{fig:part2}
% \end{figure}

% \begin{figure}[h]
%   \centering
%   \includegraphics[width=1\textwidth]{correct_analysis_case.pdf}
%   \caption{Samples of incorrect explanation.}
%   \label{fig:correct_analysis_case}
% \end{figure}
% \begin{figure}[h]
%   \centering
%   \rotatebox{270}{\includegraphics[width=1.65\textwidth]{correct_analysis_case.pdf}}
%   \caption{Samples of incorrect explanation.}
%   \label{fig:correct_analysis_case}
% \end{figure}

\subsection{Analysis of Few-Shot and Chain-of-Thought Prompts}
%我们还在answer prediction and reasoning任务上设计了few-shot prompt和chain-of-thought prompt，并在GPT模型上进行了实验。结果如表所示，可以看到使用few-shot或者chain-of-thought prompt，对于prediction任务没有太大的提升，但是在reasoning任务上均有较为明显的提升，对于GPT4模型，使用few-shot prompt，BLUE-1指标可以从0.17提升至5.95,BLUE-4指标可以从0.06提升至2.25，而使用chain-of-thought prompt，BLUE-1指标可以进一步提升至7.29。GPT35模型上也有类似的效果，使用Few-shot prompt可以使BLUE-1,BLUE-4指标可以提升至14.62和4.80，ROUGE-1, ROUGE-2, ROUGE-L分别提升至38.08, 18.35, 18.37。这是因为few-shot prompt提供了示例，因此GPT模型在生成解释的过程中，会参考示例对每一个选项进行详尽的解释，同样chain-of-though也会达到相似的效果。
In our research, we designed few-shot and chain-of-thought prompts for the answer prediction and reasoning tasks and conducted experiments on the GPT models. As shown in Table \ref{tab:few_shot_cot_prompt}, our results demonstrate that while the use of few-shot or chain-of-thought prompts did not yield significant improvements in the prediction task, but there was a notable enhancement in the reasoning task. 

Specifically, for the GPT-4 model, the utilization of few-shot prompts increased the BLUE-1 from 0.17 to 5.95, and the BLUE-4 from 0.06 to 2.25. Furthermore, incorporating chain-of-thought prompts further increased the BLUE-1 to 7.29. Similarly, positive effects were observed on the GPT-3.5 model, where few-shot prompts improved the BLUE-1 and BLUE-4 to 14.62 and 4.80, respectively. Additionally, the ROUGE-1, ROUGE-2, and ROUGE-L increased to 38.08, 18.35, and 18.37. 

These improvements can be attributed to the fact that few-shot prompts provide examples that GPT models can reference when generating detailed explanations for each option during the reasoning process. Similarly, chain-of-thought prompts can achieve similar effects, aiding in the enhancement of model performance.

\begin{table}[h]
\centering
\caption{Few-shot and chain-of-thought prompting experiment results of GPT models}
\resizebox{0.8\textwidth}{!}{%
\begin{tabular}{lccccccc}
\toprule
\multirow{2}{*}{Models} & \multicolumn{2}{c}{Prediction} & \multicolumn{5}{c}{Reasoning} \\ 
\cmidrule(rl){2-3} \cmidrule(lr){4-8}
\multicolumn{1}{c}{} & ACC & f1 & BLUE-1 & BLUE-4 & ROUGE-1 & ROUGE-2 & ROUGE-L \\ 
\midrule
GPT-4 & 61.6\%±0.1 & 61.7\%±0.1 & 0.17±0.00 & 0.06±0.00 & 29.74±0.09 & 14.84±0.04 & 11.51±0.03 \\
GPT-4\_few-shot & 62.0\%±0.4 & 61.4\%±0.5 & 5.95±0.12 & 2.25±0.07 & 37.24±0.35 & 19.23±0.26 & 17.24±0.07 \\
GPT-4\_cot & 61.6\%±0.9 & 61.4\%±0.9 & 7.29±0.71 & 2.20±0.25 & 35.85±0.78 & 16.79±0.83 & 17.18±0.30 \\
\midrule
GPT-3.5 & 46.4\%±0.0 & 46.2\%±0.1 & 3.56±0.08 & 1.49±0.06 & 33.80±0.11 & 16.39±0.05 & 14.83±0.13 \\
GPT-3.5\_few-shot & 45.3\%±0.6 & 44.9\%±0.6 & 14.62±0.16 & 4.80±0.06 & 38.08±0.44 & 18.35±0.16 & 18.37±0.29 \\
GPT-3.5\_cot & 47.9\%±0.7 & 47.7\%±0.7 & 13.47±0.52 & 3.69±0.18 & 36.47±0.42 & 16.41±0.24 & 17.82±0.31 \\ \bottomrule
\end{tabular}%
}
\label{tab:few_shot_cot_prompt}
\end{table}

\subsection{Data statistics}
Questions in CMExam have a median length of 17 (Q1: 12, Q3: 32). Regarding solution explanations, the median length is 146 tokens (Q1: 69, Q3: 247). Table \ref{tab:basic_stats} shows more basic statistics of CMExam,

\begin{table}[H]
\caption{Basic statistics of CMExam. Q: questions; E: explanations; Q1/3: the first/ third quantile.}
\centering
\resizebox{0.7\textwidth}{!}{
\begin{tabular}{lcccc}
\hline
 & {Train} & {Dev} & {Test} & {Total} \\
\hline
Question \# & 54,497 & 6,811 & 6,811 & 68,119 \\
Vocab & 4,545 & 3,620 & 3,599 & 4,629 \\
Max Q tokens & 676 & 500 & 585 & 676 \\
% Max A tokens & 5 & 5 & 5 & 5 \\
Max E tokens & 2,999 & 2,678 & 2,680 & 2,999 \\
Avg Q tokens & 29.78 & 30.07 & 32.63 & 30.83 \\
% Avg A tokens & 1.08 & 1.07 & 1.07 & 1.07 \\
Avg E tokens & 186.24 & 188.95 & 201.44 & 192.21 \\
Median (Q1, Q3) Q tokens & 17 (12, 32) & 18 (12, 32) & 18 (12, 37) & 18 (12, 32) \\
% Median (Q1, Q3) A tokens & 1 (1, 1) & 1 (1, 1) & 1 (1, 1) & 1 (1, 1) \\
Median (Q1, Q3) E tokens & 146 (69, 246) & 143 (65, 247) & 158 (80, 263) & 146 (69, 247) \\
\hline
\end{tabular}}
\label{tab:basic_stats}
\end{table}

\subsection{Guidelines for Expert-Annotation}

During the annotation phase, we invited one expert physician from the Second Affiliated Hospital of Zhejiang University and one senior doctoral student from Zhejiang University School of Medicine to carry out the annotations. The expert physician has over two years of clinical experience. The annotation guidelines have the following sections:

\begin{enumerate}
    \item Comprehensive Question Understanding: Prior to initiating the annotation process, meticulously comprehend the medical question, ensuring a holistic grasp of its context and significance.
    \item Subject Categorization: Identify the precise subject or medical field that the question pertains to, such as cardiology, pediatrics, or pathology.
    \item Principal Symptoms or Medical Conditions: Ascertain and pinpoint the primary symptoms or medical conditions expounded in the question.
    \item Examination of Pertinent Factors: Scrutinize the question for any associated factors that might be present, including the severity of the ailment, its etiology, and patient history given in the question.
    \item Examination of Pertinent Factors: Scrutinize the question for any associated factors that might be present, including the severity of the ailment, its etiology, and patient history given in the question.
    \item Appropriate Classification System Usage: Use the accurate classification system for annotation in alignment with the determined subject and symptoms. Suitable systems could encompass the 11th revision of the International Classification of Diseases (ICD-11), the Directory of Medical Institution Diagnostic and Therapeutic Categories (DMIDTC), and others.
    \item Addressing Multiple Annotations: In scenarios where the question encompasses multiple symptoms or medical conditions, opt for the most related classification for annotation.
    \item Ensuring High-Quality Annotations: Adhere to the guidelines and definitions within the chosen classification system. This diligence helps avert subjectivity and ambiguity, fostering precision in the annotations.
    \item Navigating Queries and Uncertainties: Should any doubts or uncertainties emerge during the annotation process, consult the official documents and glossaries of the chosen classification system. Engaging in discussions with professionals is also advised to achieve clarity.
    \item Resolving Discrepancies: When disagreements emerge between annotators, a collaborative discussion shall be initiated. The objective is to reach a consensus and unify the annotation decision.
\end{enumerate}


\subsection{Prompt strategies for Pre-Annotation}

During the experimental process, we indeed tried different prompts to enable GPT to better understand and complete the annotation task. The specific strategies were as follows:
\begin{enumerate}
\item Without ICD-11 Category Instructions: We did not provide detailed ICD-11 category information as instruction but directly supplied the question information and asked GPT to respond. Under this setup, a significant portion of the categories returned by GPT did not strictly belong to ICD-11 classifications, yielding unsatisfactory results.
\item Batch Processing for Cost Efficiency: Initially, we concatenated multiple questions and, through a single dialogue, had GPT return annotations for multiple questions. Under this setup, expert validation showed that the accuracy of GPT's annotations was relatively low.
\item Consistency in Formatting: When no format guidance was given, GPT's return format was inconsistent, resulting in a higher parsing cost. Hence, after multiple trials, we eventually opted for more rigorous format guidance.

Our annotation process was carried out in two stages: First, GPT conducted an initial round of pre-annotation. Subsequently, we invited an expert physician from the Second Affiliated Hospital of Zhejiang University and a doctoral student from Zhejiang University School of Medicine to annotate. The expert physician had over two years of clinical experience. In instances where there were disagreements in annotations, both annotators would discuss and eventually arrive at a consensus.
\end{enumerate}

\end{CJK*}

\end{document}